\documentclass[10pt,journal,compsoc]{IEEEtran}

\usepackage{graphicx}
\usepackage{amsfonts}
\usepackage{amsmath}
\usepackage{amssymb}
\usepackage{diagbox}
\usepackage{multirow}
\usepackage{graphics}
\usepackage{epstopdf}
\usepackage{tabularx}
\usepackage{epsfig}
\usepackage{color,subfigure}
\usepackage{times}
\usepackage{mathrsfs}
\usepackage{bm}
\usepackage{algorithm, algorithmic}
\usepackage{mathdots}
\usepackage[american]{babel}
\usepackage{microtype} 
\usepackage{array}
\usepackage{setspace}
\usepackage{threeparttable}
\usepackage{url}
\usepackage[colorlinks,linkcolor=red]{hyperref}

\ifCLASSOPTIONcompsoc

  \usepackage[nocompress]{cite}
\else
  \usepackage{cite}
\fi

\ifCLASSINFOpdf
\else
\fi

\begin{document}

\title{Unified Adversarial Patch for Visible-Infrared Cross-modal Attacks in the Physical World}

\author{Xingxing Wei$^\ddagger$, \IEEEmembership{Member,~IEEE}, Yao Huang, Yitong Sun and~Jie Yu% <-this % stops a space
\IEEEcompsocitemizethanks{
\IEEEcompsocthanksitem Xingxing Wei, Yao Huang, Yitong Sun and Jie Yu are with the Institute of Artificial Intelligence, Beihang University, No.37, Xueyuan Road, Haidian District, Beijing, 100191, P.R. China.
(E-mail: \{xxwei, y\_huang, yt\_sun, sy2106137\}@buaa.edu.cn)% <-this % stops an unwanted space
\IEEEcompsocthanksitem Xingxing Wei is the corresponding author ($\ddagger$)}
}

% \markboth{IEEE TRANSACTIONS ON PATTERN ANALYSIS AND MACHINE INTELLIGENCE}%
% {Shell \MakeLowercase{\textit{et al.}}: Bare Demo of IEEEtran.cls for Computer Society Journals}

\IEEEtitleabstractindextext{%
\begin{abstract}
Physical adversarial attacks have put a severe threat to DNN-based object detectors. To enhance security, a combination of visible and infrared sensors is deployed in various scenarios, which has proven effective in disabling existing single-modal physical attacks. To further demonstrate the potential risks in such cases, we design a unified adversarial patch that can perform cross-modal physical attacks, achieving evasion in both modalities simultaneously with a single patch. Given the different imaging mechanisms of visible and infrared sensors, our work manipulates patches' shape features, which can be captured in different modalities when they undergo changes. To deal with challenges, we propose a novel boundary-limited shape optimization approach that aims to achieve compact and smooth shapes for the adversarial patch, making it easy to implement in the physical world. And a score-aware iterative evaluation method is also introduced to balance the fooling degree between visible and infrared detectors during optimization, which guides the adversarial patch to iteratively reduce the predicted scores of the multi-modal sensors. Furthermore, we propose an Affine-Transformation-based enhancement strategy that makes the learnable shape robust to various angles, thus  mitigating the issue of shape deformation caused by different shooting angles in the real world. Our method is evaluated against several state-of-the-art object detectors, achieving an Attack Success Rate (ASR) of over 80\%. We also demonstrate the effectiveness of our approach in physical-world scenarios under various settings, including different angles, distances, postures, and scenes for both visible and infrared sensors.
\end{abstract}

}

\maketitle

\IEEEdisplaynontitleabstractindextext

\IEEEpeerreviewmaketitle

\IEEEraisesectionheading{\section{Introduction}\label{sec:introduction}}

\IEEEPARstart{W}{ith} the found vulnerability of Deep Neural Networks (DNNs) against adversarial examples \cite{szegedy2013intriguing}, DNNs' security has become a serious problem, which may cause target evasion in the physical world\cite{wei2023visually,xu2020adversarial,zhu2021fooling,zhu2022infrared}. Leveraging this property, researchers develop stealth technology to evaluate how robust DNNs are in real-world systems through physical attacks. In real life, many safety-critical tasks such as security monitoring, autonomous driving, etc., use both visible light sensors and thermal infrared sensors to perform object detection. The round-the-clock application is made possible by the combination\cite{yuan2023mathbf,wei2015structured}, which provides detailed texture information during the day under the visual modality and the target's thermal distribution at night under the infrared modality. Therefore, creating a cross-modal physical attack to trick both visible and infrared object detectors simultaneously is a critical step for robustness evaluation in such multi-modal imaging circumstances.

However, existing physical attacks are generally restricted to single modality. Some studies\cite{liang2021parallel,dongviewfool,thys2019fooling,  wei2022simultaneously,dong2023benchmarking} can circumvent detection in the optical modality, while some studies\cite{zhu2021fooling,zhu2022infrared,wei2022hotcold,xingxing2023physically} achieve evasion under infrared object detectors. Those physical attacks cannot fool multi-modal object detectors at the same time due to different imaging mechanisms. In particular, infrared sensors cannot capture the perturbations generated in the optical modality, and vice versa, the imaging of visible light domain is unaffected by changes in the object’s thermal radiation. In the digital world, there also exist some cross-modal attacks\cite{abdelfattah2021adversarial,tu2021exploring,wang2022adversarial}, but their methods mainly focus on altering the image pixels or point clouds data after sensors' imaging and ignore the imaging process of diverse mechanisms across sensors, which causes limited effectiveness in the physical world. 

    \begin{figure}[t]
    \begin{center}
       \includegraphics[width=\linewidth]{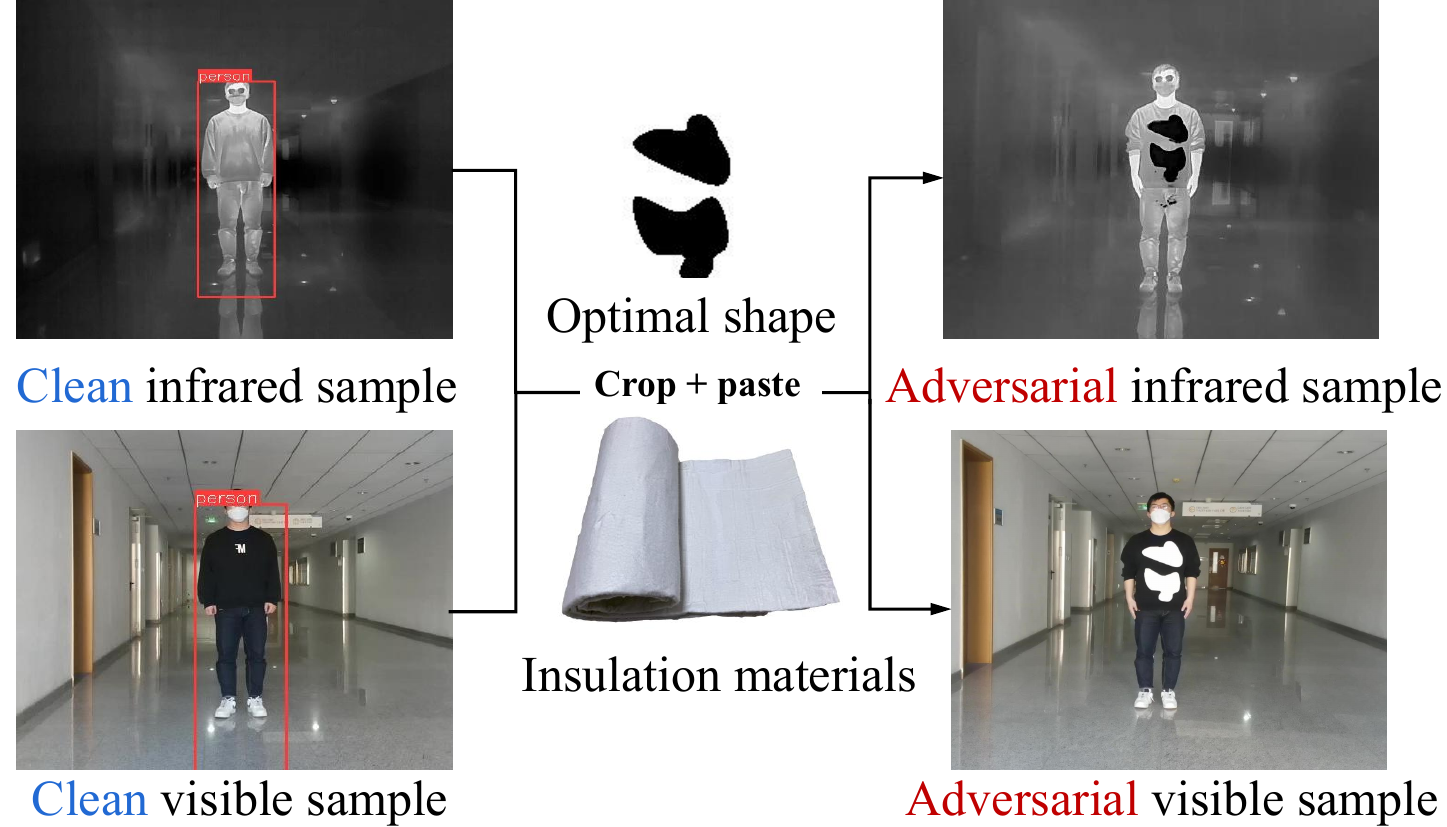}
    \end{center}
    \vspace{-0.3cm}
       \caption{The generation process for our unified cross-modal adversarial patches. We see the pedestrian cannot be detected after the patches are pasted on the pedestrian in the physical world.}
    \label{fig:pic1}
    \end{figure}

Based on the above discussions, a unified cross-modal attack in the physical world with the consideration of imaging differences is developed in this paper to fill in the gap of this field, as shown in Table \ref{tab:robustness evaluation}. In general, the root cause of limited accessibility of cross-modal physical attacks is the absence of properties that may function in several modalities. Inspired by\cite{wei2022hotcold} and \cite{chen2022shape}, we utilize adversarial patches \cite{DBLP:journals/corr/abs-1712-09665} with shape optimization to conduct attacks. The changes of patches' shape can be effectively captured by both visible and infrared sensors as a universal characteristic for diverse imaging mechanisms and thus, this shape-based method is suitable for performing cross-modal attacks. However, three core challenges still remains to meet our goal: \textbf{(1)} The current shape models are either heteromorphic \cite{chen2022shape} or change-limited \cite{wei2022hotcold}, leading to difficult physical implementation or poor attacks because of finite searching space and possible deviations. So how to design a flexible shape optimization method while keeping shapes' smoothness is the first challenge. \textbf{(2)} During the original optimization for the unified shape, the detection scores usually drop at an inconsistent rate, which may lead to significantly effective performances in one modality but fail in another. Therefore, how to balance the performance of the two object detectors from different domains is the second challenge. \textbf{(3)} When capturing the patch in the real world, a large shooting angle of the camera may lead to a shape deformation of the patch in the photo, which will degrade the performance of our shape-based physical attacks. Thus, how to optimize a robust shape to various  angles is the third challenge.  

\begin{table}[t]
\caption{Various adversarial attacks in different settings. }
  \begin{center}
   \begin{tabular}{c|c|c}
    \hline
     & Digital world   & Physical world\\
    \hline
 \multirow{2}{*}{Single-modal}    &  \cite{szegedy2014intriguing},\cite{carlini2017towards},\cite{moosavi2016deepfool},\cite{goodfellow2014explaining}, & 
 \cite{zhu2021fooling},\cite{zhu2022infrared},\cite{xingxing2023physically},\\ 
 & \cite{kurakin2016adversarial},\cite{madry2017towards}, etc. &
 \cite{sharif2016accessorize}, \cite{komkov2021advhat}, \cite{wei2022adversarial},  etc. \\   \hline
    Cross-modal  & \cite{abdelfattah2021adversarial},\cite{tu2021exploring}, etc. & Ours \\
    \hline
    \end{tabular}
    \label{tab:robustness evaluation}
    \vspace{-0.3cm}
  \end{center}
\end{table}

To address these issues, \textbf{firstly}, a novel boundary-limited shape optimization method is designed to generate compact and smooth shapes that are simple to execute in the real world, meanwhile it can provide a wider search space to flexibly find the optimal shape for a successful cross-modal attack. \textbf{Secondly}, we introduce a score-aware iterative evaluation mechanism that can direct the patch to iteratively reduce the predicted scores of the multi-modal sensors with concurrent validity, in order to balance the fooling degree of visible and infrared detectors during shape optimization process. \textbf{Thirdly}, to mitigate the issue of shape deformation due to different shooting angles, we propose an Affine-Transformation-based enhancement strategy to simulate the various patch shape's deformations along with the camera angles' change, and then search for the optimal shape that has the least drop of the attack performance  under different deformations.  We finally demonstrate the effectiveness of our cross-modal adversarial patch against visible and infrared pedestrian detectors in Figure \ref{fig:pic1}, where we only need to print and crop the simulated results in the digital world with insulation materials for patches in the physical world. The implementation is simple and convenient. The code can be found in \url{https://github.com/Aries-iai/Cross-modal_Patch_Attack}.

In summary, the contributions of this paper are as follows:
\begin{itemize}
     \item We propose a unified adversarial patch to perform cross-modal attacks. To the best of our knowledge, it is the first work to simultaneously evade visible detectors and infrared detectors in the physical world. 
     
    \item We design two novel techniques: boundary-limited shape optimization and score-aware iterative evaluation, to generate feasible patches in the digital world that can balance the multi-modal object detectors. Moreover, our framework is scalable and can be easily integrated with the patch’s position for a joint optimization.

    \item We mitigate the issue of shape deformation due to different shooting angles in the physical world and propose an Affine-Transformation-based enhancement strategy to make the learnable shape robust to various angles.

    \item We evaluate our unified cross-modal adversarial patches on the pedestrian detection task in both the digital and physical worlds. Experimental results demonstrate that our unified cross-modal adversarial patches can perform well under various angles, distances, postures, and scenes. We also apply our method to the vehicle detection task to show its good generalization ability.
    
\end{itemize}

\par This journal paper is an extended version of our ICCV paper \cite{wei2023unified}. Compared with the conference version, we have made significant improvements and extensions in this version versus the following aspects: \textbf{(1)} We further analyze the challenge of applying shape-based attacks in the physical world, and find that shape deformation will happen in the photo when meeting large shooting angles of the camera, which will degrade the performance of physical attacks. For that, we  design an Affine-Transformation-based enhancement strategy to mitigate this issue. The technical details are introduced in Section \ref{multi-angles}, and the corresponding experiments are given in Section \ref{AT-ES}.  \textbf{(2)} We show that our shape-based attack method is easy to combine the patch's position to jointly improve the attack performance, which verifies the good flexibility of our framework. The technical details are introduced in Section \ref{combined with position}, and the corresponding experiments are given in Section \ref{position optimization}. \textbf{(3)} Except for pedestrian detection,  we extend our method to attack vehicle detection task in Section \ref{PA-VD}, which demonstrates the good generalization ability of our method. \textbf{(4)} We add new experiments about parameter tuning, comparisons with SOTA methods, attacking various object detectors in Section \ref{hyt}, Section \ref{more-sota}, Section \ref{pdds}, respectively. \textbf{(5)} We polish the whole paper and add or rewrite some sections (like Sections \ref{sec:introduction}, \ref{sec:shape_wors}, \ref{sec:cross_works}, \ref{sec:experiments}, etc.) to make the paper easy to understand. We believe these modifications can significantly improve the paper's quality. 

 \par The rest of this paper is structured as follows: Section 2 reviews the related works. Section 3 describes the detailed method for our unified adversarial patch. Section 4 presents and analyzes the experimental results. Section 5 concludes the paper.

\section{Related Works}
\subsection{Adversarial Attacks in the Physical World}

Since Kurakin \emph{et al.} \cite{kurakin2016adversarial} prove the feasibility of adversarial attacks in the real world, numerous physical attack methods for various purposes have been proposed. In the visible domain, Sharif \emph{et al.} \cite{sharif2016accessorize} design adversarial glasses to deceive face recognition systems, Eykholt \emph{et al.} \cite{eykholt2018robust} create adversarial graffiti to mislead automatic driving tasks, and Xu \emph{et al.} \cite{xu2020adversarial} develop adversarial T-shirts to evade person detectors. Different from traditional L$_p$-norm based attacks which need to constrain the perturbation magnitude, these methods are based on adversarial patches \cite{brown2017adversarial}. In the infrared domain, Zhu \emph{et al.} propose adversarial bulbs \cite{zhu2021fooling} and invisible clothes \cite{zhu2022infrared} to attack infrared pedestrian detectors, respectively using extra heat sources and QR code patterns. The two methods are effective but very complex to implement in the real world. For this reason, Wei \emph{et al.} \cite{xingxing2023physically} propose a convenient adversarial infrared patch with learnable shapes and positions, it has good generalization, which can not only attack pedestrian detectors but also vehicle detectors. 

However, these methods only focus on single modality, and cannot work well for cross-modality physical attacks.

\subsection{Shape Optimization in Adversarial Attacks}
\label{sec:shape_wors}
Given the attack feasibility of early adversarial patches with fixed common shapes, some studies have looked further into patch's shape for adversarial attacks. Chen \emph{et al.} \cite{chen2022shape}, for instance, propose a deformable patch representation determined by a central point and multiple rays, making the deformable shape overly harsh and unnatural. Their work only focuses on image classification tasks and patch's content optimization is still required. Then, Wei \emph{et al.} \cite{wei2022hotcold} suggest a hotcold block based on Warming Paste and Colding Paste to attack infrared object detectors, but the deformation is limited by the manually set nine-square-grid states, which significantly decreases the search space of shape and attack performances for challenging attack goals. Besides, it relies on too much blocks, some of which are at unstable positions and may cause performance degradation. In adversarial infrared patches \cite{xingxing2023physically}, their shape modeling is essentially determined by a designed aggregation regularization. Although this method has made a good trade-off between the attack effect and shape's physical feasibility compared to the previous two methods, we still find that it overlooks the edge smoothing of infrared patches, and thus poses bad effects on physical implementation to some extent.

Unlike previous methods that use either nine-square-grid states \cite{wei2022hotcold} or polygonal shapes with a central point and rays \cite{chen2022shape}, we propose a novel approach that uses multiple anchor points and spline interpolation to generate smooth and flexible shape edges. This enables us to explore a larger search space and achieve better performance in the cross-modal physical attacks.

\subsection{Cross-modal Adversarial Attacks}
\label{sec:cross_works}
\begin{figure}[ht]
\begin{center}
\includegraphics[width=\linewidth]{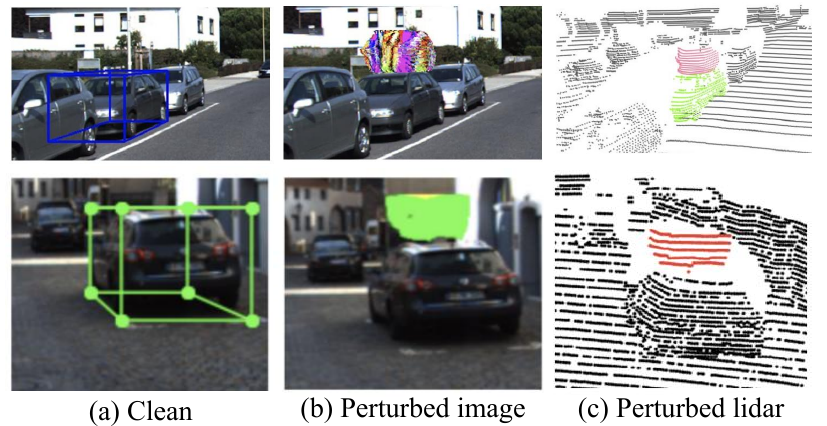}
\end{center}
   \caption{Visualization examples of two cross-modal physically realizable attacks\cite{abdelfattah2021adversarial,tu2021exploring} in RGB-LIDAR modality.}
\label{fig:contrast}
\end{figure}

Some studies have explored cross-modal attacks on multi-modal detection systems, which are mainly for autonomous vehicles based on RGB cameras and LiDAR sensors for perception, using 2D image data and 3D point clouds. For example, Abelfattah \emph{et al.}\cite{abdelfattah2021adversarial} propose a cross-modal and physically realizable attack that places an adversarial 3D object on top of a car in a 3D scene and renders it to both point clouds and RGB images using differentiable renderers, as shown in Figure \ref{fig:contrast} (top row). The object’s shape and texture are trainable parameters that can be manipulated adversarially. Tu \emph{et al.}\cite{tu2021exploring} also propose a cross-modal and physically realizable attack that inserts a textured mesh onto the car’s roof, perturbing both shape and texture information for both RGB images and point clouds, as shown in Figure \ref{fig:contrast} (bottom row). They claim that such placement can be realized in the real world.

\begin{figure*}[ht]
\begin{center}
\includegraphics[width=0.95\linewidth]{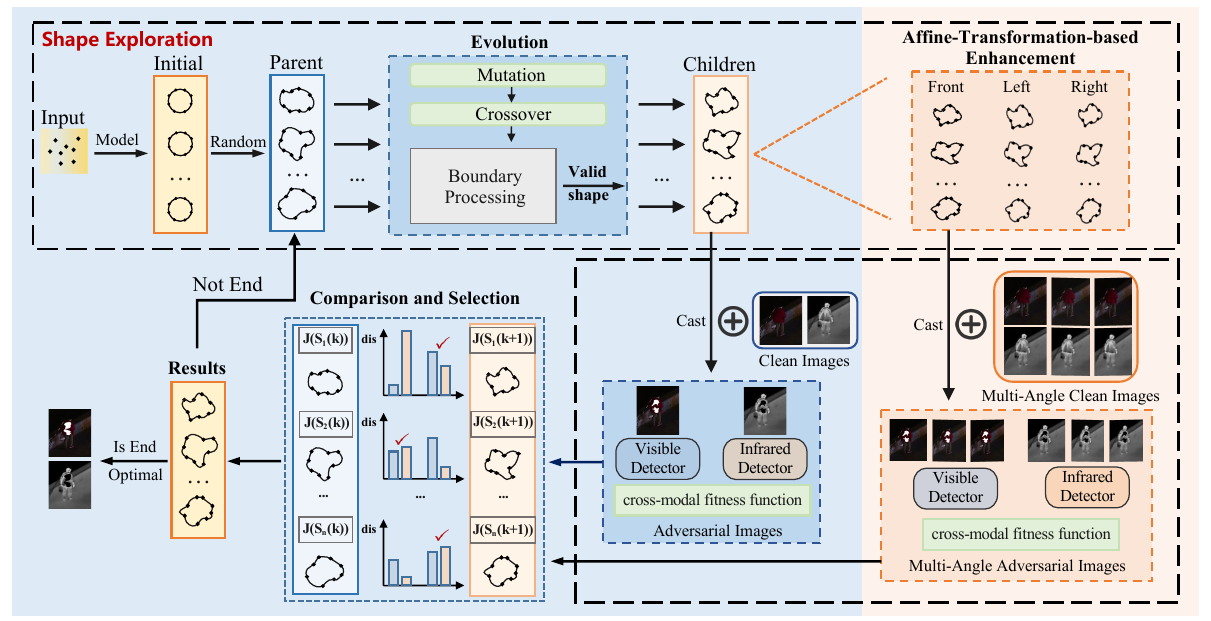}
\end{center}
\vspace{-0.5cm}
   \caption{An overview of cross-modal adversarial patches' generation based on the Differential Evolution (DE) framework. The initial population is a series of circles. Then, combining mutation, crossover and boundary processing, a child population made up of natural shapes is generated. With a special cross-modal evaluation, we compare the parent population with the child population and select better individuals, encouraging the population to become more balanced and more adversarial. It should also be mentioned that in order to ensure the robustness of angular deflection in the physical world, we design an affine-transformation-based data enhancement in the process of evaluation. Finally, the optimal individual will be printed.}
\label{fig:framework}
\end{figure*}

But the existing methods only claim to be ``physically realizable", which have not been tested in the real world. They only demonstrate their methods in a simulated 3D environment, and rely on the consistency of adversarial 3D objects across modalities to prove their physical realizability. However, such adversarial 3D objects are difficult to produce and impractical to place over a vehicle. Therefore, ``physically realizable" only means theoretically feasible, not actually applicable.

In contrast to the existing methods that only ensure ``physically realizable" attacks in simulation, we present \textbf{the first cross-modal physical attack} that actually works on the object in the real world. Moreover, we target visible-infrared cross-modal attacks, while \cite{abdelfattah2021adversarial} and \cite{tu2021exploring} target visible-LiDAR cross-modal attacks, which is another major difference between our work and theirs. To the best of our knowledge, none of the previous methods can simultaneously attack both visible and infrared modalities in reality.

\section{Methodology}
In this section, we choose pedestrian detection as the target task to introduce the details of our method.%, and first introduce the deformable patch representation by centripetal catmull–rom spline, then detail the proposed shape-based Differential Evolution algorithm and finally introduce the calculation method of the cross-modal fitness function in the process of shape optimization.
\subsection{Problem Formulation}

In the pedestrian detection task, given a clean visible image $x_{vis}$ and a clean infrared image $x_{inf}$, the goal of a unified cross-modal adversarial attack is to make the visible detector $f_{vis}(\cdot)$ and infrared detector $f_{inf}(\cdot)$ 
simultaneously unable to detect the pedestrian in the perturbed visible image $x^{adv}_{vis}$ and perturbed infrared image $x^{adv}_{inf}$. The formulation can be expressed as follows:
\begin{equation}
    max(f_{vis}(x^{adv}_{vis}), f_{inf}(x^{adv}_{inf})) < thre
    \label{attack}
\end{equation}
where $f_{vis}(x^{adv}_{vis})$ and $f_{inf}(x^{adv}_{inf})$ represent the confidence scores of detected pedestrians in the visible modality and infrared modality, and $thre$ is a pre-defined threshold.

The perturbed visible image $x^{adv}_{vis}$ and the perturbed infrared image $x^{adv}_{inf}$ with the unified adversarial patch can be generated as  Eq.(\ref{eq:x_vis_adv}) and Eq.(\ref{eq:x_inf_adv}):
\begin{equation} 
    x^{adv}_{vis} = x_{vis}\odot (1 - M) + \hat{x}_{vis}\odot M
    \label{eq:x_vis_adv}
\end{equation}
\begin{equation}
    x^{adv}_{inf} = x_{inf}\odot (1 - M) + \hat{x}_{inf}\odot M
    \label{eq:x_inf_adv}
\end{equation}
where $\odot$ is Hadamard product, $M \in \{0,1\}^{h\times w}$ is a mask matrix used to constrain the shape and location of the cross-modal patches on the target object, $\hat{x}_{vis}\in R^{h\times w}$ denotes a cover image used to manipulate $x^{adv}_{vis}$, $\hat{x}_{inf}\in R^{h\times w}$ denotes a cover image used to manipulate $x^{adv}_{inf}$. The values of these two cover images are obtained by the visible sensor and infrared sensor shooting the unified insulation material in the physical world. The unified patches can be described as all the regions where $M_{ij} = 1$. 

In the real application, we use the aerogel material to implement our unified adversarial patch. When infrared sensor shoots, it will show good insulation effects, and thus can change the thermal distribution of the target pedestrian. When visible sensor shoots, its color is white and can show a difference with the pedestrian. Based on this, our method  optimizes a cross-modal $M$ to learn an adversarial shape, finally working well in both modalities. 

\subsection{Shape Representation}
To model a shape, we need to determine the shape representation. For that, we first define some points as the basic elements. Then we connect these anchor points to construct the contour, representing our shape. Here are the details.

\textbf{Multi-anchor Representation:}
Unlike Chen \emph{et al.} \cite{chen2022shape} using a central point and corresponding rays to form a polygon contour, we only use points, distributing multiple anchor points on the patch contour. Then, we can directly adapt the patch contour's shape by changing the coordinates of the points. Owing to the design of multi-anchor representation, we can have more flexible shape variations and a broader search space.

\begin{figure}[ht]
\begin{center}
\includegraphics[width=\linewidth]{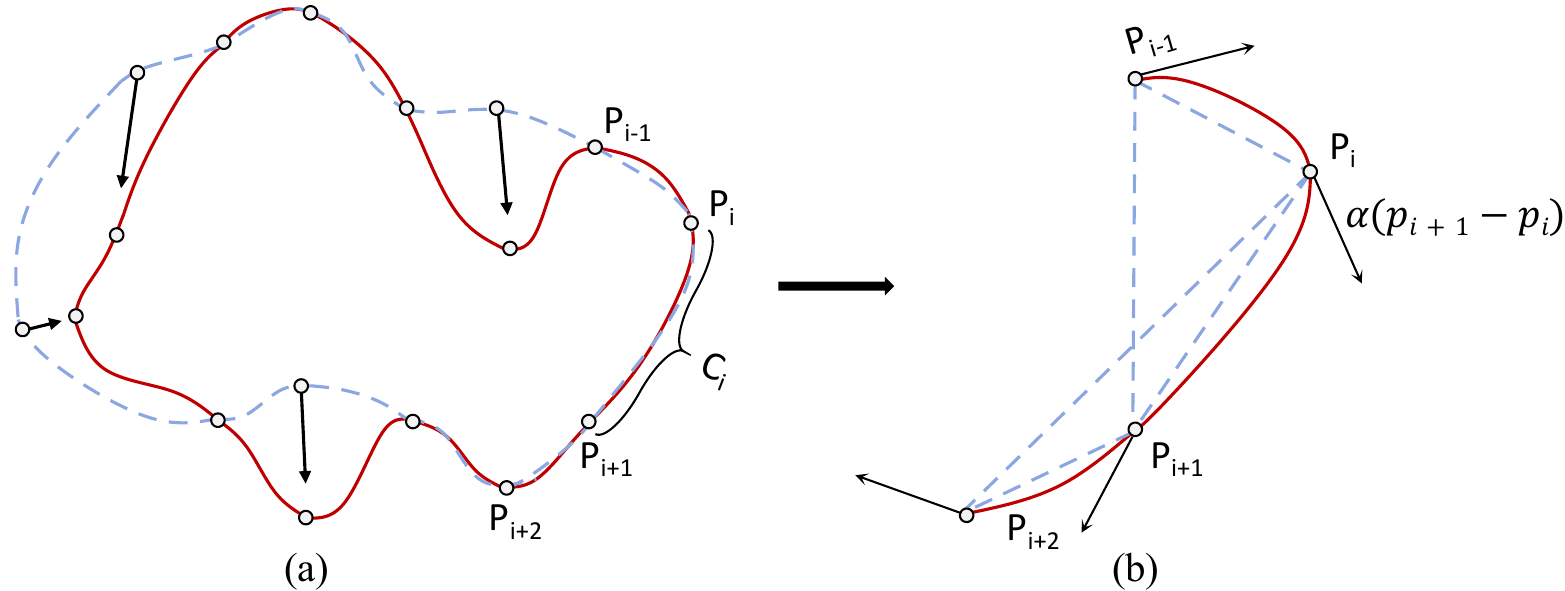}
\end{center}
\vspace{-0.3cm}
\caption{Subfigure (a) shows the process of moving the anchor points to change the shape, where the dotted line denotes the initialized location, and arrows show the changed direction. Subfigure (b) is an example of the curve segment $C_{i}$'s connection via Catmull-Rom spline interpolation. }
\label{fig:patch_contour}
\end{figure}

We illustrate this process in Figure \ref{fig:patch_contour} (a), where points in the dotted line denote the initialized location. The arrows show the changed direction for some points. We can adjust the coordinates of the points to control the movement of each point and thus, the direction of our points' movement will not be limited.

\textbf{Smooth Spline Connection:}
To ensure shapes' naturalness, we use centripetal Catmull–Rom spline \cite{twigg2003catmull} to connect such anchor points. The process of spline can be formulated as follows:

As shown in Figure \ref{fig:patch_contour} (b), to generate a curve segment $C_i$ between $P_i$ and $P_{i+1}$, we will use four anchor points $P_{i-1},P_{i},P_{i+1},P_{i+2}$ and a centripetal Catmull-Rom spline function $CCRS(\cdot)$ \cite{twigg2003catmull}. With the centripetal Catmull-Rom spline function, first, we can ensure that a loop or self-intersection doesn't exist within a curve segment. Second, a cusp will never occur within a curve segment. Finally, it can also follow the anchor points more tightly. $C_i$ can be formulated as follows:
\begin{equation}
    C_{i} = CCRS(P_{i-1},P_{i},P_{i+1},P_{i+2})
\label{C_i}
\end{equation}

When $n$ curve segments $C_0,C_1,\cdots,C_{n-1}$ are combined, the patch contour $M_{con}$ can be written as:
\begin{equation}
    M_{con} = \{C_i|0\leq i\leq n-1\}
    \label{Mcon}
\end{equation}

With $M_{con}$ being closed, we can easily obtain $M$:
\begin{equation}
M(x)= \begin{cases}
1,&\text{$x$ \textit{inside} $M_{con}$} \\
0,&\text{$x$ \textit{out of} $M_{con}$}
\end{cases} 
    \label{M}
\end{equation}
After Eq.(\ref{M}), we can represent the patch's shape. The detailed formula for this process can be found in \textit{Supplementary Material}.

\label{deformable}
\begin{figure*}[ht]
\begin{center}
\includegraphics[width=\linewidth]{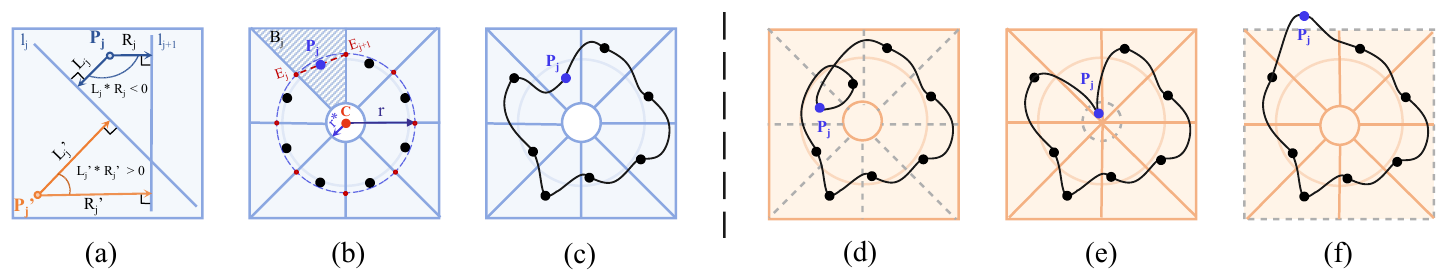}
\end{center}
\vspace{-0.5cm}
   \caption{The solid blue and orange boxes represent the effective area of human body and the dashed line in orange area means that the part is missing. The (a) expresses that the product of the directed distances from the point in the sharp-angled sector to the two side lines is less than 0 and if not, the product is more than 0. The (b) is our initial state and the (c) is an effective deformation. The (d) shows the patch contour will cross if we don't fix its relative order. The (e) represents that connecting points will be trapped into a restricted space if we don't set an inner circle. The (f) expresses that the patch contour will go beyond the human body if we don't define an outer boundary.}
\label{fig:boundary}
\end{figure*}
\subsection{Shape Optimization for Cross-modal Attacks}

In practice, the details of $f_{vis}(\cdot)$ and $f_{inf}(\cdot)$ are usually unknown to the adversary, so it is intractable to optimize the anchor points via a gradient-based optimization method. Considering this background, we carry out a score-based black-box attack by querying the object detector to obtain the confidence scores of detected pedestrians. From the above, we formulate the whole unified cross-modal attacks into the Differential Evolution (DE) framework\cite{price2006differential}.
% Thus, we formulate our unified cross-modal attacks into the Differential Evolution (DE) framework.

\subsubsection{Formulation Overview Using DE}
The Differential Evolution (DE) consists of four parts: starting from an initial  population, using the crossover and mutation to generate the offspring population, making the fittest survive according to the fitness function, and finding the appropriate solution in the iterative evolution process.

In our case, a population represents the anchor points $\{P_j|j=1,...,n\}$. Given the population size $Q$, the $k$-th generation solutions $S(k)$ is represented as:
\begin{equation}
    S(k)=\{S_{i}(k)|\theta^{L}_{j}\leq S_{ij}(k)\leq \theta_{j}^{U}, 1\leq i\leq Q, 1\leq j\leq n\}
    \label{S(k)}
\end{equation}
where $S_{i}(k)$ is the $i$-th patch's shape, and $S_{ij}(k)$ represents the $j$-th anchor point of $S_{i}(k)$ in the $k$-th generation. $\theta^{L}_{j}$ and $\theta^{U}_{j}$ together make up the feasible region $B_{j}$, which is the moving range of the $j$-th anchor point in each patch shape.

In the $k+1$ generation of DE, the solution $S(k+1)$ is achieved via crossover, mutation, and selection based on $S(k)$. A fitness function is applied on $S_{i}(k)$ to evaluate its attack effectiveness. During this process, the fitness function only utilizes the confidence scores of object detectors, therefore, we can conduct score-based black-box attacks.  We will give the detailed definition  of feasible region in Section \ref{boundary}, and fitness function in Section \ref{fitness}.

\subsubsection{Boundary-limited Shape Optimization}
\label{boundary}

Obviously, if there is no boundary restriction for the anchor points during the crossover and mutation,  some bad situations will occur during shape optimization, as shown in Figure \ref{fig:boundary} (d, e, f),  such as boundary line crossings (d), anchor points stuck in narrow places (e), and anchor points outside the human body region (f). To address these issues, we propose a novel boundary-limited shape optimization, which not only provides sufficient transformation space, but also ensures the effectiveness of each shape. 

As shown in Figure \ref{fig:boundary} (b), the shaded part $B_{j}$ is exactly the feasible region of the anchor point $P_{j}$. Specifically, the two adjacent equidistant lines, the inner circle's edge and the outer border together make up of its boundary. Figure \ref{fig:boundary} (c) presents us with an example of effective deformation. Next, we will give how to construct the boundary and the method of judging whether a point is inside or not.

We first divide a circle with a given radius $r$ and a circle center $C$ into $n$ sharp-angled sectors by $n$ equidistant lines $\{l_{j}|j=1,\cdots,n\}$. $n$ equal points $\{E_{j}|j=1,\cdots,n\}$ are the intersection of equidistant lines and the circle, and as shown in Figure \ref{fig:boundary} (b), the anchor point $P_{j}$ is the midpoint of $E_{j}$ and $E_{j+1}$, which can be formulated as follows:
\begin{equation}
    P_{j} = \frac{E_{j}+E_{j+1}}{2},\quad j=1,\cdots,n
\label{equal point}
\end{equation}

For the anchor point $P_{j}$, it lies in the sharp-angled sector region $B_{j}$ wrapped by two adjacent lines $l_{j}$ and $l_{j+1}$. $L_{j}$ and $R_{j}$ represent the directed distance from the anchor point $P_{j}$ to the line $l_{j}$ and $l_{j+1}$ shown in the Figure \ref{fig:boundary} (a). Therefore, with the line function $l_{j}(x,y)=0$ and $l_{j+1}(x,y)=0$, we can formulate $L_{j}$ and $R_{j}$ as follows:
\begin{equation}
    L_{j} = \frac{l_{j}(x_j,y_j)}{D_{j}},\quad R_{j} = \frac{l_{j+1}(x_j,y_{j})}{D_{j+1}}
\end{equation}
where $D_{j}$ and $D_{j+1}$ are the denominators of the point-to-line distance formula.
 
From the Figure \ref{fig:boundary} (b), we can vividly find that since the anchor point $P_{j}$'s feasible region is a sharp-angled sector, if $P_{j}$ is inside its own region, $L_{j}*R_{j}<0$ for the angle between them is obtuse, while if $P_{j}$ is outside its own region or exactly on the boundary line,  $L_{j}*R_{j}\geq0$. With $P_{j}$ inside its own region, we can effectively avoid boundary line crossings in Figure \ref{fig:boundary} (d) caused by spline interpolation's need for a given order. 

Then, to prevent points from falling into narrow places as Figure \ref{fig:boundary} (e), we set an inner circle with a given radius $r^{*}$ inside the initial circle, and anchor points are not allowed to move into the inner circle. Specifically, we can judge by the distance $r_{j}$ from the anchor point $P_j$ to the center $C$. When $r_{j}>r^{*}$, the anchor point $P_{j}$ will not be inside the inner circle. The $r_{j}$ is computed as follows:  
\begin{equation}
    r_j = \sqrt{(x_{j}-x_{C})^2+(y_{j}-y_{C})^2}
\end{equation}

After that, we still need to ensure that the generated patch is not outside the effective area of the human body as shown in Figure \ref{fig:boundary} (f), so we scale the detection box output in a certain proportion as the outer border. The region $\mathcal{O}$ inside the outer border can be represented as follows:
\begin{equation}
    \mathcal{O} = \{(x,y)|x_{l}\leq x\leq x_{r}, y_{d}\leq x\leq y_{u}\}
\end{equation}
where $x_{l},x_{r},y_{u},y_{d}$ are outer border’s vertex coordinates.

Finally, we combine the above limits as follows:
\begin{equation}
    \rho_{j} = (L_{j}*R_{j}<0)\ \&\ (r_{j}>r^{*})\ \&\ (P_{j}\in \mathcal{O})
\label{rho}
\end{equation}
where $\rho_{j} = 1$ means that the anchor point $P_{j}$ is inside the feasible region $B_{j}$ and $\rho_{j} = 0$ means the anchor point $P_{j}$ is outside $B_{j}$.

\subsubsection{Score-aware Iterative Evaluation}
\label{fitness}
In cross-modal attacks, a good attack effect of a single modality is ineffective, while it is common to have unbalanced attack effects in the two modalities. If this situation is not improved, it may give the attacker false signals about the progress of the attack. Thus, to balance the fooling degree between visible detector and infrared detector during the optimization process, we propose a score-aware iterative evaluation, guiding the adversarial patch to iteratively reduce predicted scores of the multi-modal sensors.

For the sake of simplicity of expression, here we denote $S_{i}(k)$ as $s$. To evaluate the fitness value of $s$, we first use methods in Section \ref{deformable} to transform $s$ into a patch mask $M_{s}$, then $x^{adv}_{vis,s}$, $x^{adv}_{inf,s}$ are produced based on Eq.(\ref{eq:x_vis_adv}), Eq.(\ref{eq:x_inf_adv}). The fitness funtion $J(s)$ can be formed as follows:
\begin{equation}
    J(s)=e^{\lambda*min(dis(x^{adv}_{vis,s}),dis(x^{adv}_{inf,s}))}
        \label{J(s)}
\end{equation}
where $\lambda$ is a weighted factor, $dis(x^{adv}_{vis,s}),dis(x^{adv}_{inf,s})$ reflect the current progress towards the success of attack
(the larger, the closer to success). $dis(x^{adv}_{vis,s}),dis(x^{adv}_{inf,s})$ can be formalized as:
\begin{equation}
    dis(x^{adv}_{vis,s})=\frac{f_{vis}(x_{vis})-f_{vis}(x^{adv}_{vis,s})}{f_{vis}(x_{vis})-thre}
        \label{dis_vis}
\end{equation}
\begin{equation}
    dis(x^{adv}_{inf,s})=\frac{f_{inf}(x_{inf})-f_{inf}(x^{adv}_{inf,s})}{f_{inf}(x_{inf})-thre}
        \label{dis_inf}
\end{equation}
where $f_{vis}(x_{vis})$ is the confidence score of $x_{vis}$ in the visible pedestrian detector, $f_{inf}(x_{inf})$ is the confidence score of $x_{inf}$ in the infrared pedestrian detector. $f_{vis}(x^{adv}_{vis,s})$ and $f_{inf}(x^{adv}_{inf,s})$ are similar to $f_{vis}(x_{vis})$ and $f_{inf}(x_{inf})$.
% \begin{figure}[ht]
%\begin{center}
%\includegraphics[width=\linewidth]{iccv2023AuthorKit/fitness.pdf}
%\end{center}
%   \caption{Example of a short caption, which should be centered.}
%\label{fig:fitness}
%\end{figure}

From Eq.(\ref{dis_vis}) and Eq.(\ref{dis_inf}), we can easily know $dis(x^{adv}_{vis,s})$ and $dis(x^{adv}_{inf,s})$ measure the progress to success of the cross-modal patch attack in the visible and infrared modality respectively. Both of them can help patches evolve in the corresponding modality. However, if we only make use of a single one, it will be certain to cause unbalanced phenomena, not an effective cross-modal attack. To solve this issue, we use Eq.(\ref{J(s)}) to combine $dis(x^{adv}_{vis,s})$ and $dis(x^{adv}_{inf,s})$. Based on our $J(\cdot)$, the good performance only in a single modality will not achieve a high value of fitness because we take the worse one of $dis(x^{adv}_{vis,s})$ and $dis(x^{adv}_{inf,s})$ as a standard. Additionally, considering the difference of attack difficulty in the initial stage and later stage, we use $e^{(\cdot)}$ instead of a linear function. Based on the above settings, we will eventually encourage it to iteratively evolve in the direction of reducing confidence scores as much as possible while maintaining the balance of cross-modalities. 

\subsubsection{Combined with Position Optimization}
\label{combined with position}
In the previous section of shape representation, a patch's shape is determined by $n$ anchor points, which can be directly adapted by changing coordinates of those points, while the initial coordinates and feasible regions of the $n$ anchor points still depend on a pre-set central point and in this way, it will determine the final pasting position of the whole patch. However, according to Wei \emph{et al.} \cite{wei2022simultaneously}, the optimized patch with the fixed position tends to be the best only in the current local part, and there may exist a more effective region in the global. Thus, it is also necessary to verify the position factor to our unified adversarial patch.

In our case, since position is also one of the major parameters on the patch attack, it is natural to think of optimizing anchor points' positions along with the initial central point, achieving simultaneous optimization of patch's shape and position. Different from Wei \emph{et al.} \cite{wei2022hotcold}'s method, which performs adjustment by gradient, we only need to add the horizontal and vertical coordinates of the central point to the set of the $n$ point coordinates of the characterized individual. Taking  the individual $s$ as an example, we denote $s=\{P_j|j=1,...,n\}$ in the previous definition, then we combine the central point $(C_x,C_y)$ with the point set to represent the new individual $s'$. At this moment, $s'$ is denoted as follows:
\begin{equation}
    s' = \{C_x,C_y, P_{1},\cdots, P_{n}\}
\end{equation}

Finally, we successfully take the exploration of patch's position into account and achieve joint optimization under the cross-modal DE algorithm. Designed comparison experiments prove that adding the attribute ``position" into the optimization process can truly improve the final attack performance and the specific results are shown in Section \ref{position optimization}. 

The overall algorithms for cross-modal attacks and unified patches' generation are summarized in Algorithm \ref{alg:456}, \ref{alg:123}, and the whole illustration for this framework is given in Figure \ref{fig:framework}.
\begin{algorithm}[h]  
    \renewcommand{\algorithmicrequire}{\textbf{Input:}}
    \renewcommand{\algorithmicensure}{\textbf{Output:}}
    \caption{Shape Optimization for Cross-modal Attacks} 
    \begin{algorithmic}[1]   
        \REQUIRE {Clean visible image $\bm{x}_{vis}$, clean infrared image $\bm{x}_{inf}$, the fitness function $J(\cdot)$, the population size $Q$, the max number of iterations $T$}
        \ENSURE visible adversarial example $x^{adv}_{vis}$ and infrared adversarial example $x^{adv}_{inf}$  
        \STATE Initialize a collection of shapes $S(0)$
        \FOR{$k = 0$ to $T-1$}  
            \STATE{Sort $\bm{S}(k)$ in descending order according to $J(S(k))$} 
            \IF{$\bm{S_0(k)}$ makes the attack successful} \STATE $stop=k$; break; 
            \ENDIF 
            \STATE Generate $S(k+1)$ based on crossover and  mutation.
            \STATE Limit boundaries of $S(k+1)$ according to Eq.(\ref{rho})
        \FOR{$i=1$ to $Q$}
            \STATE Evaluate $S_{i}(k)$ and $S_{i}(k+1)$ according to Eq.(\ref{J(s)})
              \STATE{$S_{i}(k+1)\leftarrow$ the better one in $S_{i}(k+1)$ and $S_{i}(k)$}
        \ENDFOR  
        \ENDFOR
        \STATE{Sort $\bm{S}(stop)$ in descending order according to $J(S(k+1))$} 
        \STATE Choose $S_{0}(stop)$ as the final individual from $\bm{S}(stop)$ 
        \STATE Generate unified patch $M$ with $S_{0}(stop)$ by Algorithm \ref{alg:123}
        \STATE Obtain adversarial examples with $M$ according to Eq.(\ref{eq:x_vis_adv}),(\ref{eq:x_inf_adv})
        \STATE{\small\textbf{return} $x^{adv}_{vis}$, $x^{adv}_{inf}$}
  \end{algorithmic}  
  \label{alg:456}
\end{algorithm}

\begin{algorithm}[h]  
    \renewcommand{\algorithmicrequire}{\textbf{Input:}}
    \renewcommand{\algorithmicensure}{\textbf{Output:}}
    \caption{Generate Unified Patch $M$ from Anchor Points} 
    \begin{algorithmic}[1]   
        \REQUIRE {Individual $s$, anchor points number $n$, the initial central points $C^{0}_{x},C^{0}_{y}$}
        \ENSURE Unified Adversarial Patch $M$ 
        \STATE  Obtain new central points $C_{x},C_{y}\leftarrow s$
        \STATE Obtain new anchor points $P_{1},\cdots,P_{n}\leftarrow s$
        \FOR{$i=1$ to $n$}
            \STATE Adapt anchor points' positions according to central points\\
            $P_{i}\leftarrow P_{i} + [C_{x},C_{y}] - [C^{0}_{x},C^{0}_{y}]$
        \ENDFOR
                \FOR{$i=1$ to $n$}
            \STATE Obtain curve segment $C_i$ with $\{P_{i-1},\cdots, P_{i+2}\}$ by Eq.(\ref{C_i})
        \ENDFOR
        \STATE{$M_{con}\leftarrow$ combine $\{C_{i=1,\cdots,n}\}$ according to Eq.(\ref{Mcon})}
        \STATE{$M\leftarrow$ fill the $M_{con}$ according to Eq.(\ref{M})}
        \STATE{\small\textbf{return} unified patch $M$}
  \end{algorithmic}  
  \label{alg:123}
\end{algorithm}
\vspace{-0.5cm}
\begin{figure*}[htp]
\begin{center}
\includegraphics[width=\linewidth]{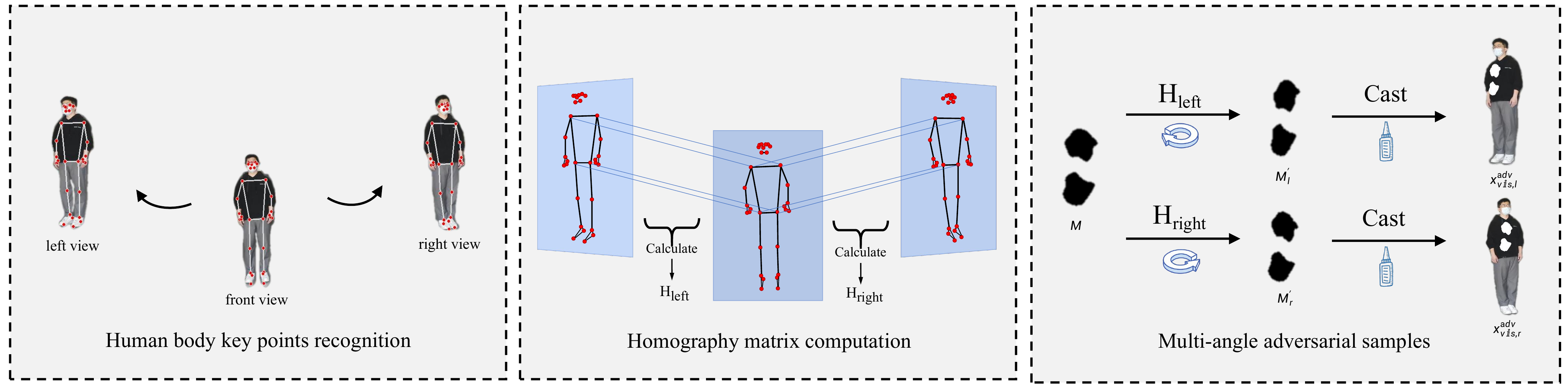}
\end{center}
\vspace{-0.5cm}
\caption{The  process of our proposed Affine-Transformation-based enhancement strategy, which contains three steps. Details are in the texts.}
\label{fig:multi-angle}
\end{figure*}

\subsection{Transfer to the Physical World}
\subsubsection{Affine-Transformation-based Enhancement Strategy}
\label{multi-angles}
Considering that when camera shoots the patch with a large shooting angle, the patch's shape will show a deformation, we need to ensure the versatility of the unified adversarial patch for multiple angles. In other words, our patch should explore the shape with good attack retention under some deformations. Towards this goal, we combine perspective transformation with human pose estimation to obtain objects with patches at different angles, and then utilize such multi-angle data to optimize our patch. 

Specifically, in the first step, to perform a more accurate perspective transformation, we need to find relatively fixed reference points, which can help us compute the better homography matrix used in the transformation. Fortunately, due to the objects being pedestrians, keypoints of human body can greatly meet our requirements. So, we take human pose estimation\cite{bazarevsky2020device} to extract such keypoints. Specifically, this method first trains a face detector to localize the human body based on the observation that the strongest signal of the neural network at the torso position is the face. Next, it predicts the midpoint of a person's hip, the radius of the outer circle of the whole person, and the tilt angle of the line connecting the midpoint of the shoulder and hip to locate 33 keypoints from the nose to the left foot, which well represents a human body structure.

After obtaining the source keypoints $(x_{i}, y_{i})_{i=1,\cdots,33}$ and the target keypoints $(x_{i}', y_{i}')_{i=1,\cdots,33}$,  we could utilize such point pairs to estimate the homography matrix $H$. However, if not all of the point pairs (source keypoints, target keypoints) fit the rigid perspective transformation, which means that there exist some outliers, this initial estimate will be poor. So, we use the RANSAC \cite{fischler1981random} method to select the best subset from the point pairs for the initial estimation. To be detailed, the method RANSAC tries many different random subsets of the corresponding point pairs (four pairs each), estimates the homography matrix using this subset and then computes the quality of the computed homography, which is the number of inliers. The criteria for  outliers are as follows:
\begin{equation}
    ||(
    \begin{array}{c}
        x'   \\
        y' 
    \end{array}
    )-
    T(H*(
    \begin{array}{c}
         x  \\
         y 
    \end{array}))
    || > ransacthre
\end{equation}
where $T(\cdot)$ is a function to convert points from homogeneous coordinates back to Euclidean coordinates and  $ransacthre$ is a pre-set threshold for distinguishing outliers and inliers in the RANSAC method \cite{fischler1981random}.

With a good initial estimation, we then approximate the $H$ by minimizing the back-projection error rate. $H$ can be formulated as follows:
\begin{equation}
H=
\left[
\begin{array}{ccc}
h_{11} & h_{12} & h_{13} \\
h_{21} & h_{22} & h_{23} \\
h_{31} & h_{32} & h_{33}
\end{array}
\right]
\end{equation}

Then, the back-projection error is computed like below:
\begin{equation}
\begin{split}
        \sum_{i}(x_{i}'-\frac{h_{11}x_{i}+h_{12}y_{i}+h_{13}}{h_{31}x_{i}+h_{32}y_{i}+h_{33}})^{2}\\
    + (y_{i}'-\frac{h_{21}x_{i}+h_{22}y_{i}+h_{23}}{h_{31}x_{i}+h_{32}y_{i}+h_{33}})^{2}
\end{split}
\end{equation}

Next, we can use such an optimal homography matrix $H$ to transform a frontal mask $M$ into an $M'$ at different angles as:
\begin{equation}
M'=HM    
\end{equation}

Finally, according to Eq.(\ref{eq:x_vis_adv}) and Eq.(\ref{eq:x_inf_adv}), adversarial samples with patches under different angles can be generated. Taking the adversarial samples $x^{adv}_{vis,l}, x^{adv}_{inf,l}$ in a left-skewed view as an example, they can be represented as follows:
\begin{equation} 
    x^{adv}_{vis,l} = x_{vis,l}\odot (1 - M'_{l}) + \hat{x}_{vis,l}\odot M'_{l}
    \label{eq:x_vis_adv_left}
\end{equation}
\begin{equation}
    x^{adv}_{inf,l} = x_{inf,l}\odot (1 - M'_{l}) + \hat{x}_{inf,l}\odot M'_{l}
    \label{eq:x_inf_adv_left}
\end{equation}
where $M'_{l}$ is a transformed mask at a left-skewed angle. $x^{adv}_{vis,r}, x^{adv}_{inf,r}$'s generation is similar to $x^{adv}_{vis,l}, x^{adv}_{inf,l}$.

After obtaining new multi-angle adversarial samples, we can design a joint fitness function $J'(s)$ like Eq.(\ref{J(s)}) to enhance our unified patches' multi-angle attack performance in the physical world. With the individual $s$, the joint fitness function $J'(s)$ can be formulated as follows:
\begin{equation}
J'(s)=e^{\lambda*min(D)}
\end{equation}
where $D$ is a set of the current progress towards the attack success in multiple angles. With $x^{adv}_{vis,s,l}, x^{adv}_{inf,s,l}, x^{adv}_{vis,s,r}, x^{adv}_{inf,s,r}$ representing adversarial samples generated based on the individual $s$, $D$ can be represented as follows:
\begin{equation}
\begin{split}
        D=\{dis(x^{adv}_{vis,s}), dis(x^{adv}_{inf,s}), dis(x^{adv}_{vis,s,l}),\\ dis(x^{adv}_{inf,s,l}), dis(x^{adv}_{vis,s,r}), dis(x^{adv}_{inf,s,r})\}
\end{split}
\end{equation}
% 用这些样本去联合优化的公式
% 公式13
% \begin{equation}
% \begin{split}
% J'(s)=e^{\lambda*min(dis(x^{adv}_{vis}),dis(x^{adv}_{inf}),dis(x^{adv}_{vis,l}),dis(x^{adv}_{inf,l}),dis(x^{adv}_{vis,r}),dis(x^{adv}_{inf,r}))} 
% \end{split}

%         \label{J'(s)}
% \end{equation}
For simplicity, we here only add two additional angles, actually, multiple different angles can be considered to augment the set $D$.

\subsubsection{Physical Implementation}
\label{physical}
\begin{figure}[ht]
\begin{center}
\includegraphics[width=\linewidth]{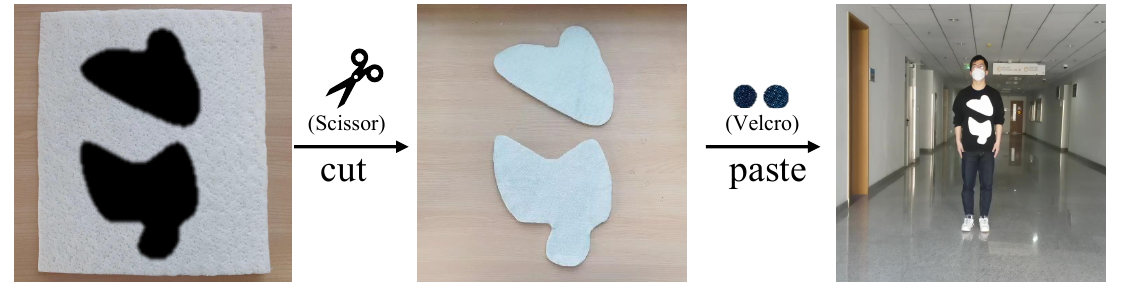}
\end{center}
\vspace{-0.5cm}
  \caption{The physical implementation process of transforming digital cross-modal patches into physical patches in the real world.}
\label{fig:implementation}
\end{figure}

After obtaining the optimal shape, we start to transform algorithm-generated digital patches into physical patches with the aerogel material. Specifically, we first scale the obtained patch according to its size in the real world and print it. Then, we use scissors to equally cut out the optimal shape on the material. And in the last, we use the velcro sticker to fix it at corresponding position of the human body for performing a physical attack. The whole process is demonstrated in Figure \ref{fig:implementation}.

\section{Experiments}
\label{sec:experiments}
\subsection{Simulation of Physical Attacks}
\noindent\textbf{Dataset:} We conduct experiments using the LLVIP dataset \cite{jia2021llvip} to simulate physical attacks. The LLVIP dataset provides perfectly synchronized images in both visible and infrared modalities, and we customize the parts containing pedestrians in the images, following the approach used in \cite{zhu2021fooling} and \cite{zhu2022infrared}. The dataset consists of 3,784 images in the training set and 1,220 images in the test set. We select 120 images from the test set that the target model could recognize with high accuracy as the final samples to be attacked, resulting in a baseline clean Average Precision (AP) of 100\%.

\noindent\textbf{Target detector:} To evaluate the effectiveness of our proposed approach for pedestrian detection, we select several mainstream detectors, including YOLOv3 \cite{redmon2018yolov3}, YOLOv5, YOLOv7\cite{wang2022yolov7}, Faster RCNN \cite{ren2015faster}, EfficientDet\cite{tan2020efficientdet}, and RetinaNet\cite{lin2017focal}. For each detector, we use the officially pre-trained weights as the initialized weights and then retrain the model on the training dataset. These models are then used as the target models in our attack process.

% YOLOv5, YOLOv7\cite{https://doi.org/10.48550/arxiv.2207.02696}, Faster RCNN\cite{ren2015faster}, SSD\cite{liu2016ssd}, and EfficientDet because we want to validate our methodology both horizontally between one-stage models and two-stage models as well as vertically from the same series of models at various times. 

\noindent{\textbf{Metrics}}: We use Attack Success Rate (ASR) and Average Precision drop (AP drop) as the metrics to evaluate the performance of our physical attacks. To highlight the effectiveness of simultaneously attacking two modalities, we adopt a unique cross-modal ASR, which measures the ratio of successfully attacked images under both visible and infrared modalities out of all the test images. The AP drop metric is used to show the variation in Average Precision before and after the attacks.

\noindent\textbf{Implementation:}  As we use the DE algorithm, we set the number of the initial population as $30$, and the epochs of evolution as $200$. All hyperparameters are verified on the validation set.
\subsubsection{Hyperparameter Tuning}
\label{hyt}
In this section, we tune hyperparameters against YOLOv3 model.

\begin{figure}[h]
\begin{center}
\includegraphics[width=0.9\linewidth]{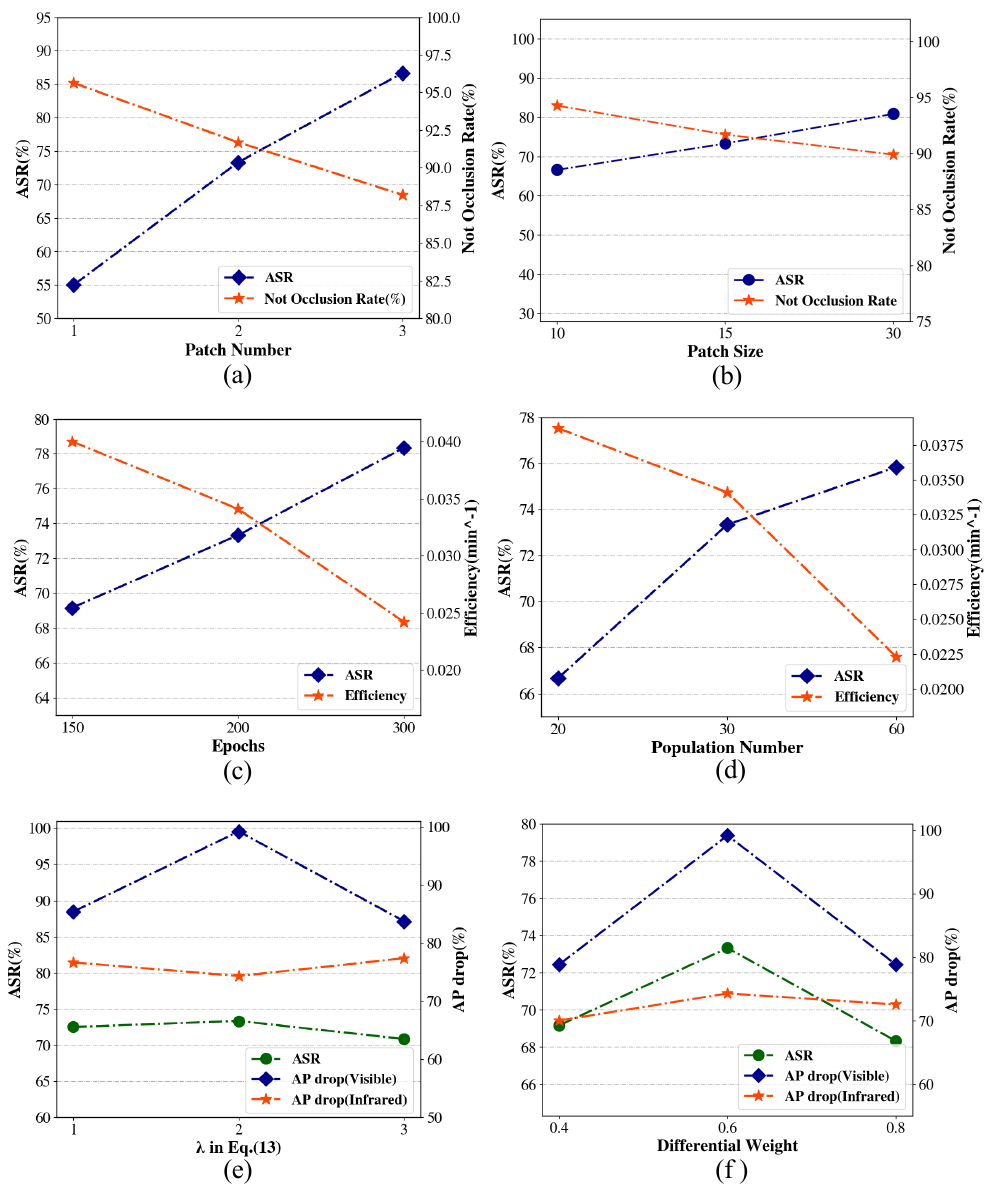}
\end{center}
\vspace{-0.3cm}
  \caption{Parameters tuning in our method. We here show the experimental results for six parameters, which are patch number, patch size, $\lambda$ in Eq.(13), epochs, population number and differential weight. Please see the text for details. }
\label{fig:hyperparameter}
\end{figure}

\noindent\textbf{Patch number.} As one of the most directly adjustable hyperparameters, patch number may have an impact on the attack performance. Therefore, we need to explore whether and how it is related to the attack performance. Figure \ref{fig:hyperparameter} (a) vividly shows that as the number of patches rises, both ASR and AP drop increase with our expectation. However, considering that our attack method has achieved acceptable results with two patches, and three patches would bring more occlusion, we finally decide to choose the patch number 2 as the main evaluation option.

\begin{figure}[!h]
    \centering
    \includegraphics[width=0.8\linewidth]{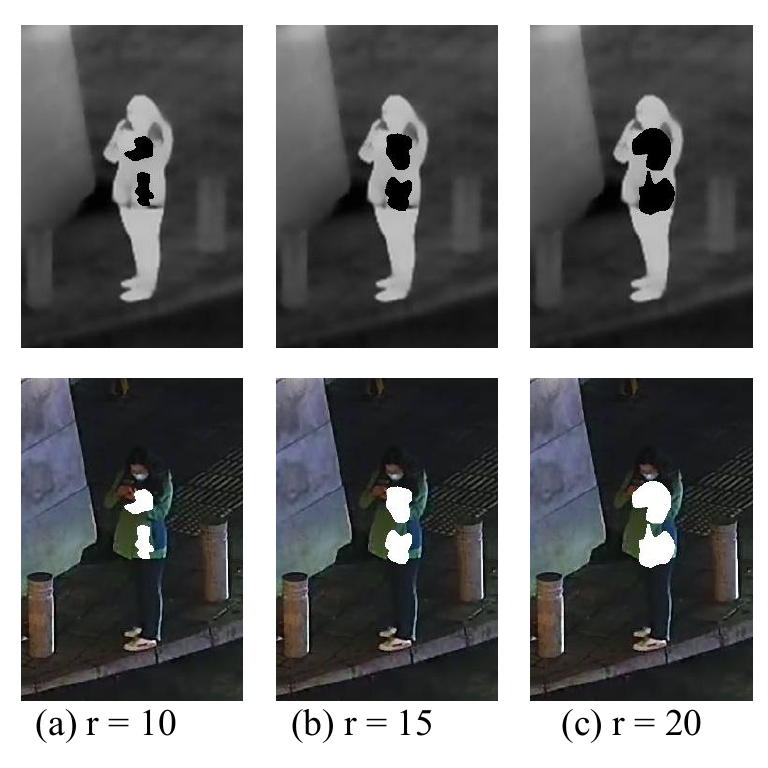}
    \caption{Visual examples of unified patches with different patch sizes.}
    \label{fig:example}
\end{figure}
\noindent\textbf{Patch size.} In our method, the radius $r$ of the initial circle affects the size of the patch to some extent. To explore the impact of patch size on attack performance, we set 10, 15 and 30 for $r$. The corresponding quantitative results are shown in Figure \ref{fig:hyperparameter} (b), where we can see that though the ASR goes up as the patch size increases, our method can still achieve a competitive result with a small patch area, like an ASR of 66.67\% with the radius $r=10$. Additionally, we measure the percentage of patches in the pedestrian area i.e. ``Occlusion Rate". The higher the occlusion rate, the easier it is to cover non-effective areas, such as the human head region, as seen in Figure \ref{fig:example} (c). Therefore, to get a trade-off between attack performance and feasibility, we finally choose a radius of $r=15$. Some examples of different patch sizes are shown in Figure \ref{fig:example}. 

% \begin{table}[!h]\centering
% \caption{The ASR(\%) and AP drop(\%) with different patch sizes.}
% {\begin{tabular}{c | c | c | c}
% \hline
%  Patch size       &  10 & 15 &  20  \\ \hline
% ASR     &     66.67\%        &     73.33\%        &      80.83\%          \\    \hline
% Occlusion Rate & 5.74\%  & 8.31\%  &   11.09\% \\\hline
% \end{tabular}}
% \label{patch size}
% \end{table}
\noindent \textbf{Epochs.} The epochs represent the maximum number of evolution to a single image. Thus, to explore the sensitivity of our method to time, we set epochs as 150, 200, 300. As Figure \ref{fig:hyperparameter} (c) shows, though  increasing evolution times can help unified adversarial patches improve themselves, its time efficiency (i.e. the countdown of time cost) has rapidly decreased. For example, when we compare the  phase of [200, 300] with the phase of [150, 200], we use nearly two times extra epochs but only obtain a similar increase in ASR. Therefore, to make a trade-off between the effect and the time efficiency, we finally determine the number of epochs as 200.

\noindent \textbf{Population Number.} The number of populations affects the population diversity. A small number of populations means that the populations do not differ much and are easy to fall into the local optimal solution, but this does not mean that a larger number of populations is better. Actually, the larger the number of populations, the more difficult it is to converge and the lower the time efficiency. From Figure \ref{fig:hyperparameter} (d), we can know that when the population number increases from $20$ to $30$, the attack performance has an improvement of 6.66\%, while the improvement decreases to 2.50\% when the population increases from $30$ to $60$. Therefore, considering the insignificant growth and low time efficiency after $30$, we decide the population number to be $30$.

\noindent\textbf{Hyperparameter $\lambda$.} The $\lambda$ in Eq.(\ref{J(s)}) affects the variability of the final fitness of the individual, the larger the $\lambda$, the greater the difference between different individuals, but this does not mean that the greater the difference, the better the attack effects. As shown in Figure \ref{fig:hyperparameter} (e), the attack effectiveness of cross-modal patches increases when $\lambda$ equals from 1 to 2, but drops dramatically when $\lambda$ equals 3. As a result, we set $\lambda=2$ to obtain the optimal effect.
% \begin{figure}[h]
% \begin{center}
% \includegraphics[width=\linewidth]{figure_zx.pdf}
% \end{center}
% \caption{The ASR(\%) and AP drop(\%) of unified adversarial patches
% with different patch numbers and hyperparameter $\lambda$.}
% \label{fig:patch_number}
% \end{figure}

% \subsubsection{Hyperparameters in Differential Evolution}

% \begin{table}[!h]\centering
% \caption{The ASR(\%) and AP drop(\%) with different epochs.}
% {\begin{tabular}{c | c | c | c}
% \hline
% Epochs        &  150 & 200 &  300  \\ \hline
% ASR     &    69.17\%      &      \textbf{73.33\%}       &     78.33\%              \\ \hline
% AP drop (Visible) &   78.05\%             &    \textbf{99.19\%}         &     86.18\%             \\ 
% AP drop (Infrared) &  71.85\%    &   \textbf{74.31\%}    &   80.33\%    \\
% \hline
% \end{tabular}}
% \label{epoch}
% \end{table}

% \begin{table}[h]\centering
% \caption{The ASR(\%) and AP drop(\%) with different population number.}
% {\begin{tabular}{c | c | c | c}
% \hline
% Population number        &  20 & 30 &  60  \\ \hline
% ASR     &        66.67\%        &      \textbf{73.33\%}       & 75.83\%                   \\ \hline
% AP drop (Visible) &        78.86\%        &      \textbf{99.19\%}       &       86.18\%           \\ 
% AP drop (Infrared) &  71.15\%    &   \textbf{74.31\%}    & 77.37\%      \\
% \hline
% \end{tabular}}
% \label{population number}
% \end{table}

\noindent \textbf{Differential Weight $\beta$.}
In the process of Differential Evolution, we use crossover and mutation between random individuals and inbreeding of superior individuals to generate candidate populations $Cr(k)$. The $Cr(k)$ can be formulated as follows:
\begin{equation}
Cr_{i}(x)= clip(S_{\gamma_{3}}(k)+\beta(S_{\gamma_1}(k)-S_{\gamma_2}(k)))
    \label{Cr(k)}
\end{equation}
where $Cr_{i}(k)$ is the $i$-th individual in the $k$-th candidate population. $\gamma_{1},\gamma_{2},\gamma_{3}$ are random numbers $\in [1,n]$. and $\gamma_{1}\neq\gamma_{2}\neq\gamma_3$. $\beta$ is the differential weight and $clip(\cdot)$ is a clipping operation to keep individuals within the range.

The differential weight $\beta$ represents the degree of population variation and also affects the convergence speed, the larger the $\beta$ the slower the convergence speed.
Therefore, when given a fixed number of epochs, we need to adjust the $\beta$ to ensure the attack performance. From Figure \ref{fig:hyperparameter} (f), we can find that the attack effectiveness of cross-modal patches increases when $\beta$ equals from 0.4 to 0.6, but drops dramatically when $\beta$ equals 0.8, which may be accounted for lack of enough epochs to achieve the convergence. As a result, we set $\beta=0.6$ to obtain the optimal effect.
% \begin{table}[!h]\centering
% \caption{The ASR(\%) and AP drop(\%) with different $\beta$.}
% {\begin{tabular}{c | c | c | c}
% \hline
% Differential weight        &  0.4 & 0.6 & 0.8  \\ \hline
% ASR     &        69.17\%        &   \textbf{73.33\%}          &       68.33\%            \\ \hline
% AP drop (Visible) &     78.86\%           &    \textbf{99.19\%}         &    78.86\%              \\ 
% AP drop (Infrared) &  70.04\%    &  \textbf{74.31\%}     &  72.56\%     \\
% \hline
% \end{tabular}}
% \label{alpha}
% \end{table}

\subsubsection{Effects of Boundary-limited Shape Optimization}
\label{optimized shapes}
 \begin{figure}[h]
\begin{center}
\includegraphics[width=0.95\linewidth]{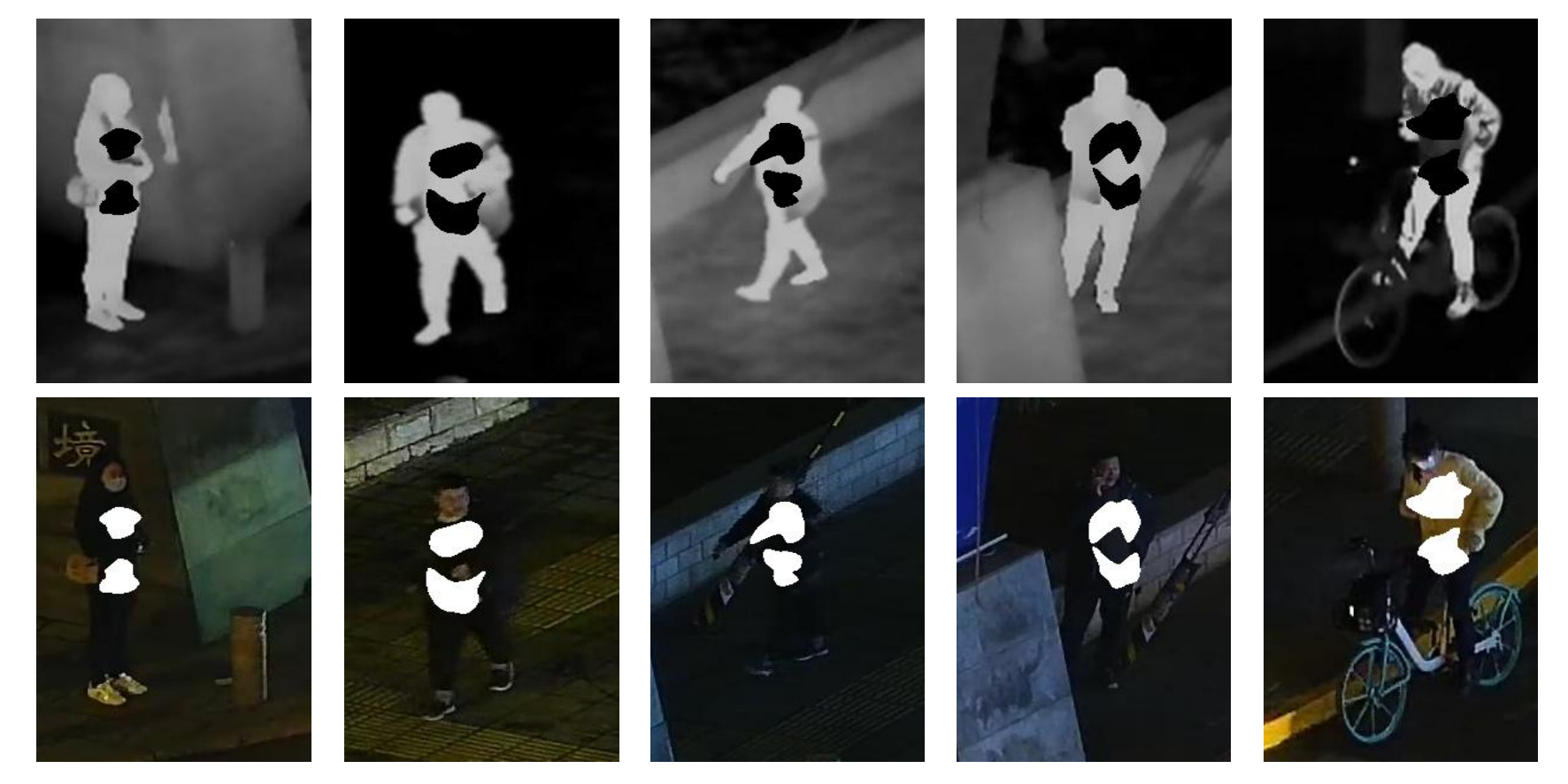}
\end{center}
   \caption{Visual examples of adversarial samples with unified adversarial patches in the digital world.}
\label{fig:digital_attack}
\end{figure}
\begin{figure}[!h]
\begin{center}
\includegraphics[width=0.95\linewidth]{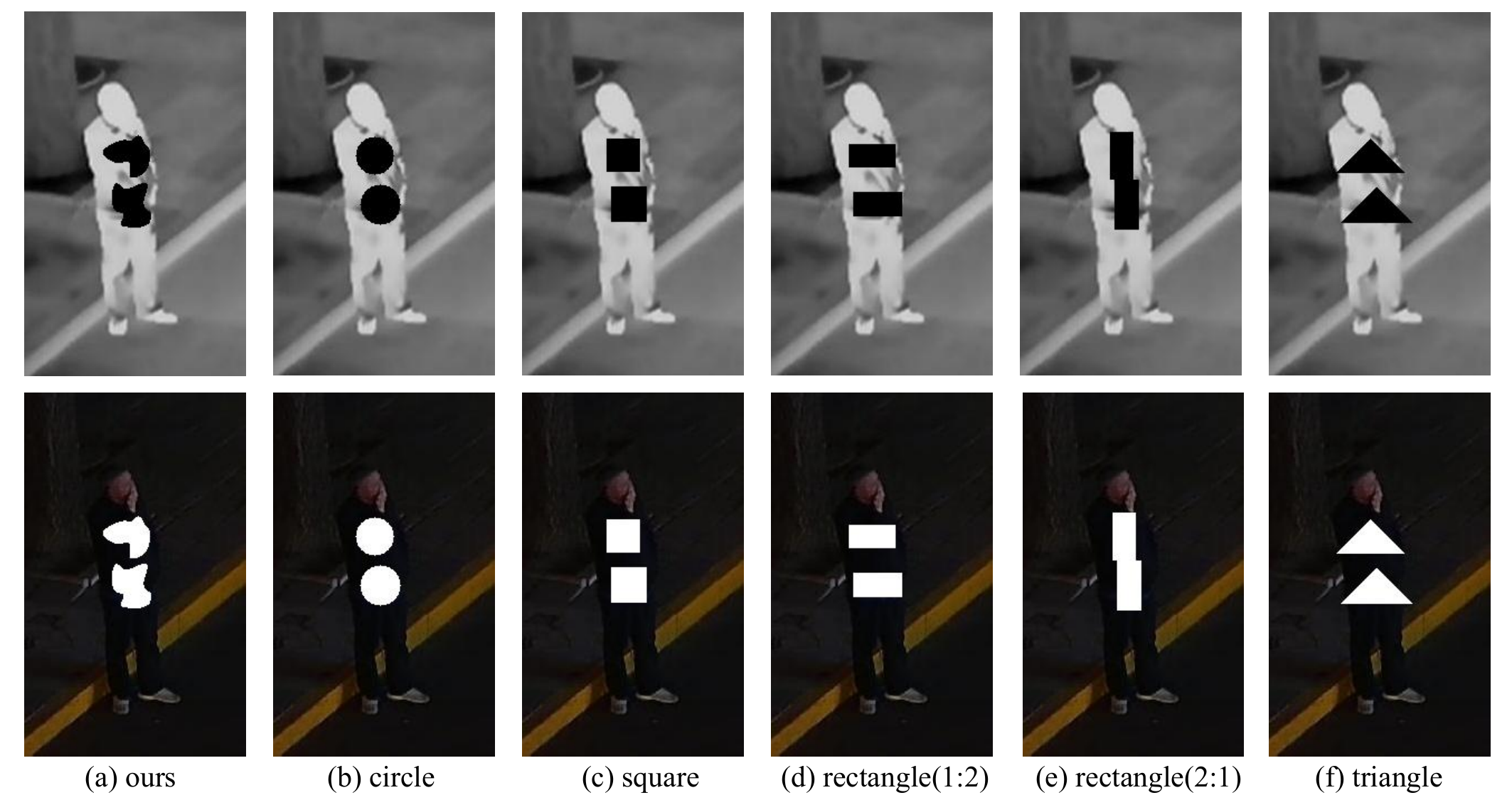}
\end{center}
\vspace{-0.3cm}
   \caption{Comparison between our optimal shapes and basic shapes.}
\label{fig:basic_shape}
\end{figure}

Here, we provide the ablation study to investigate the outcomes of our optimized shapes. Specifically, we first generate cross-modal adversarial patches to achieve the optimal shapes for a given pedestrian. Some examples of the optimal shapes in the digital world are shown in Figure \ref{fig:digital_attack}. Then, we fix the patches' locations and sizes while changing their shapes into basic shapes. These basic shapes can be circles, squares, rectangles and triangles, which are presented at Figure \ref{fig:basic_shape}. Then we compute the ASR and AP drop under these five shapes. This setting verifies the effectiveness of our optimal shapes on the attacks. The results are listed in Table \ref{shapes}, where we see that compared with our optimal shapes, basic shapes barely work with an average ASR of 5.23\%, an average AP drop of 20.73\% in the visible modality and 11.65\% in the infrared modality. 
\vspace{-0.3cm}
\begin{table}[h]\centering
\caption{Ablation study for unified adversarial patches' shapes against YOLOv3.}
\resizebox{\linewidth}{!}{
\begin{tabular}{c | c | c | c }
% \toprule[1.1pt] 
\hline
Shape     &  ASR & AP drop(Visible)  & AP drop(Infrared)\\ \hline
% \midrule
Circle    &   15.00\%    & 26.83\%  & 42.28\% \\   
Square &   2.50\%   &  22.76\%   &  4.07\% \\   
Rectangle(1:2) &  0.00\%  & 17.07\%  & 0.81\%  \\  
Rectangle(2:1) & 3.33\%  & 31.71\%  & 8.94\%  \\
Triangle & 5.83\% & 26.02\%  & 13.82\% \\
Average & 5.23\% & 20.73\%  &  11.65\% \\
Ours &  \textbf{73.33\%} & \textbf{99.19\%}  & \textbf{74.31\%} \\ \hline
% \bottomrule[1.1pt]  
\end{tabular}}
\label{shapes}
\end{table}
\vspace{-0.3cm}

\subsubsection{Effects of Score-aware Iterative  Evaluation}
\label{exp in cross-modal fitness function}
To verify the impact of the Cross-modal Fitness Function, we design a set of comparative experiments. As mentioned in Section \ref{fitness}, our method can combine the visible modality and infrared modality into a whole, which helps balance attack performances between different modalities. Here, we use a direct sum of Eq.(\ref{dis_vis}) and Eq.(\ref{dis_inf}) instead of Eq.(\ref{J(s)}) as the not-combined fitness function.
\vspace{-0.3cm}
\begin{table}[!h]
\centering
   \caption{Ablation study for score-aware iterative fitness function against YOLOv3.}
\begin{tabular}{c |c |c}
  \hline  
% \toprule[1.1pt]
Fitness Function        &  Ours & Direct Sum  \\   \hline  
% \midrule
ASR     &       \textbf{ 73.33\% }   &  50.83\%      \\   \hline  
AP drop (Visible)  &  \textbf{     99.19\% }   &  85.37\%  \\   
AP drop (Infrared)   & \textbf{74.31\%}  &  66.99\% \\ \hline  
 % \bottomrule[1.1pt]
\end{tabular} 
\label{combine}
\end{table}

From Table \ref{combine}, we can find that although a direct sum fitness function can have an AP decrease of 85.37\% in the visible modality and 66.99\% in the infrared modality, it suffers an obvious decline in ASR under a cross-modal standard.

\begin{figure}[!h]
    \centering
    \includegraphics[width=\linewidth]{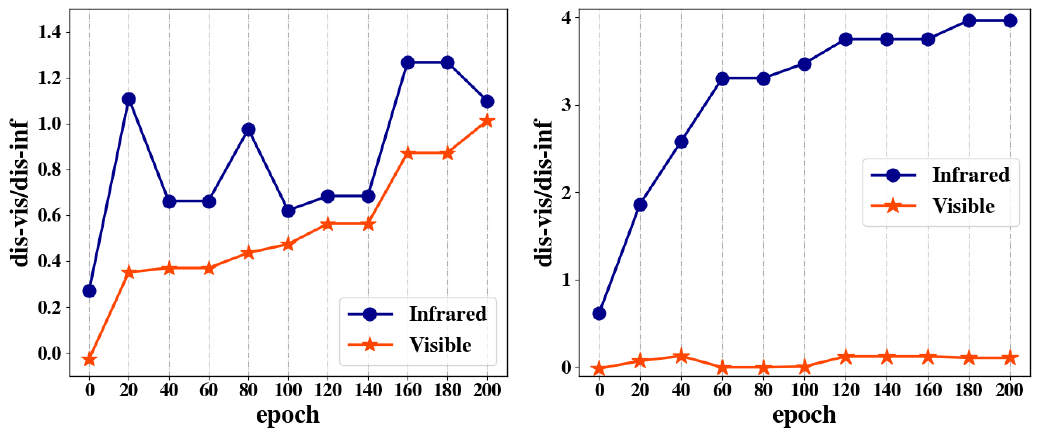}
    \caption{Visualization of specific optimization processes in score-aware iterative function and a simple sum function. $dis$-vis/$dis$-inf in y-axis denotes the value of $dis_{vis}$ and $dis_{inf}$ respectively.}
    \label{fig:zx}
\end{figure}

Then, to further demonstrate the effectiveness of our method, we visualize their specific optimization processes. As shown in Figure \ref{fig:zx}, with $dis_{vis}$ and $dis_{inf}$ representing the current progress towards the success of attack in the corresponding modality
(Eq.(\ref{dis_vis})-Eq.(\ref{dis_inf}) in the paper, the larger, the closer to success), our method balances the differences between two modalities and achieves simultaneous progress for both, whereas a simple sum tends to focus on a single easy-to-attack modality, such as the infrared modality in Figure \ref{fig:zx} (b).

\subsubsection{Effects of Integrating with Position}
\label{position optimization}
As position may also be an important attribute to the attack performance, we conduct a comparison experiment to quantify the effect of combining the shape with the position. The results listed in Table \ref{tab:compare position} show that, the ASR of joint optimization of shape and position has reached 83.33\% with an increase of 10\%, which confirms the necessity to consider the  position attribute.

\begin{table}[!h]\centering
\caption{The comparison with different optimization attributes against YOLOv3}
{\begin{tabular}{c | c | c }
\hline
 Optimization attributes       &  Shape &  Shape + Position  \\ \hline
ASR     &      73.33\%       &       \textbf{83.33\%}                       \\ \hline
AP drop (Visible) &     99.19\%          &     \textbf{94.31\%}                       \\ 
AP drop (Infrared) &   74.31\%   &   \textbf{88.30\%}         \\
\hline
\end{tabular}}
\label{tab:compare position}
\end{table}

\subsubsection{Compared with Other Shape Optimization Methods}
\label{more-sota}

\begin{figure}[!h]
    \centering
    \includegraphics[width=0.95\linewidth]{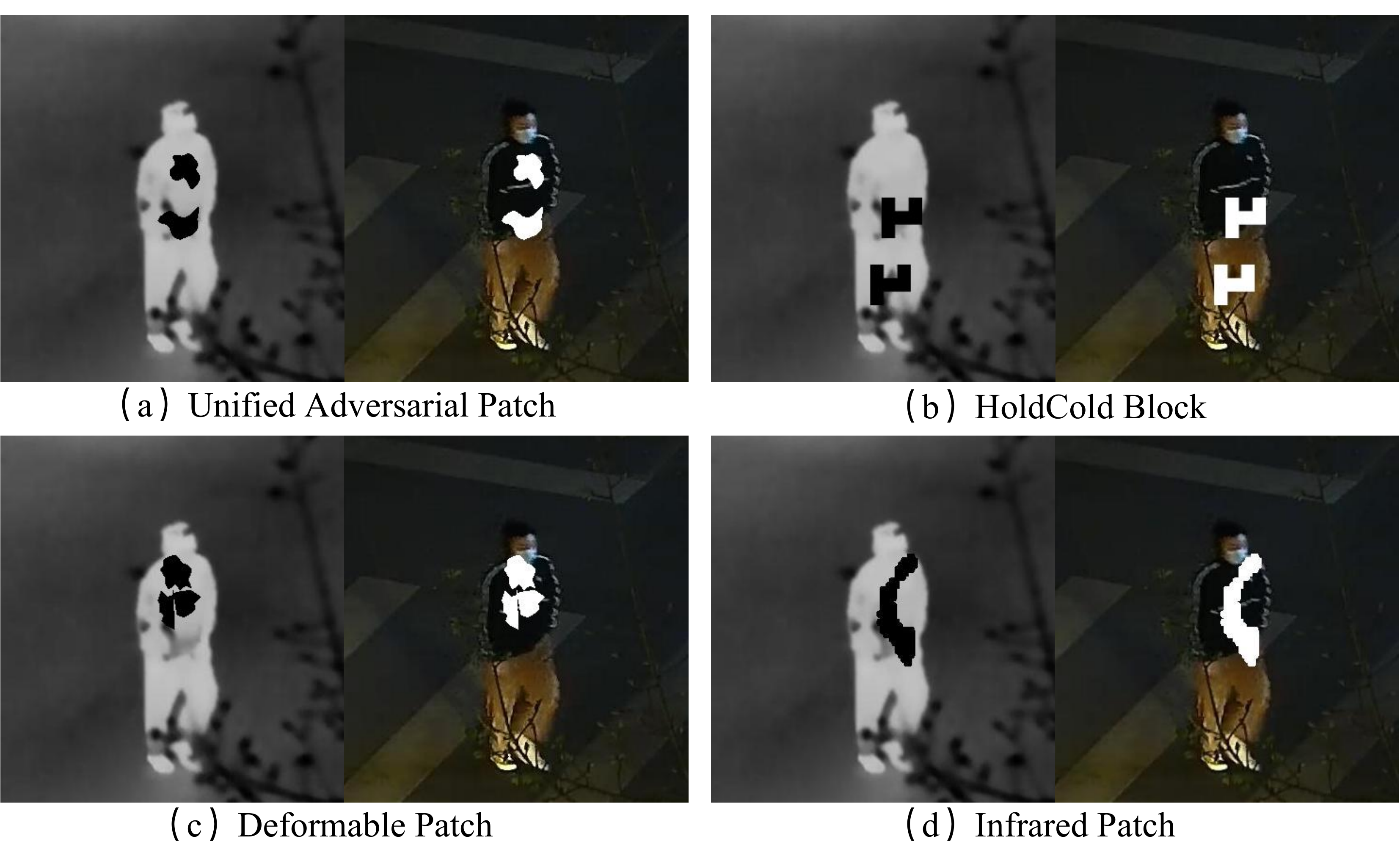}
    \caption{Visual examples generated by our unified patches, hotcold blocks and deformable patches, respectively.}
    \label{fig:compare}
\end{figure}

Wei \emph{et al.}\cite{wei2022hotcold} propose a shape optimization of utilizing nine-square-grid shapes to attack infrared detectors (called as hotcold block). Since their method is also a black-box attack, we can easily combine Wei \emph{et al.}\cite{wei2022hotcold}'s  shape modeling manner with our score-aware iterative function to conduct the comparison. We ensure to use the same patch number and patch size for these two methods. From Table \ref{tab:compare}, we can see that the hotcold block only has an ASR of 30.83\%, which supports our belief that the search space of such nine-square-grid shapes is greatly
limited. Moreover, as Figure \ref{fig:compare} (b) shows, some hotcold block's optimized positions may not be possible for patches to fix at, causing the difficulty and instability of physical implementation.

\begin{table}[!h]\centering
\caption{ Comparisons between various shape-based methods against YOLOv3}
\resizebox{\linewidth}{!}
{\begin{tabular}{c | c | c | c | c }
\hline
 Shape Methods       &  Ours & \cite{wei2022hotcold} & \cite{chen2022shape}  & \cite{xingxing2023physically}\\ \hline
ASR     &      \textbf{83.33\%}          &       30.83\%  &    56.67\%    &58.33\%              \\ \hline
AP drop (Visible) &      \textbf{94.31\%}          &     51.54\%  &  82.11\%  &  64.48\%                   \\ 
AP drop (Infrared) &   \textbf{88.30\%}   &   45.53\%   &   58.33\% &63.91\%  \\
\hline
\end{tabular}}
\label{tab:compare}
\end{table}

We also compare our method with deformable shape \cite{chen2022shape}.We give a visualization comparison for the optimized shapes between \cite{chen2022shape} and ours in Figure \ref{fig:compare} , where we can see that the deformable patches are heteromorphic as shown in Figure \ref{fig:compare} (c), not easy to implement accompanied by potentially greater implementation errors in the physical world. In contrast, our shape in Figure \ref{fig:compare} (a) is more natural, and thus is easy-to-implement to attach on the pedestrian to achieve an effective physical attack. Besides, \cite{chen2022shape} aims to attack image classifiers in the visible domain, while ours aims to attack object detectors in both the visible and infrared domains, which is another difference between the two methods. 

\begin{figure*}[htp]
\begin{center}
\includegraphics[width=\linewidth]{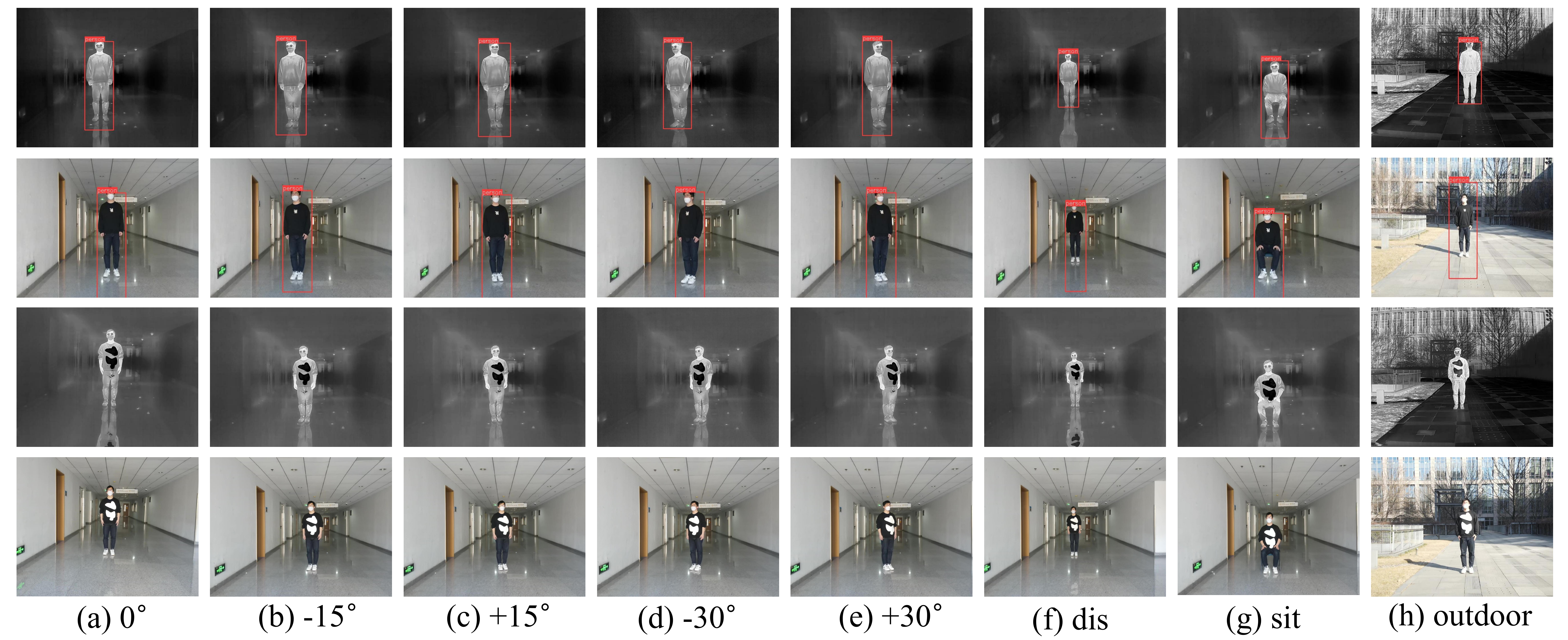}
\end{center}
\caption{Visual examples of physical attacks with unified adversarial patches under various angles, distances, postures, and scenes. }
\label{fig:physical}
% \vspace{-0.5cm}
\end{figure*}

Moreover, we utilize the latest shape optimization method in Wei \emph{et al.}\cite{xingxing2023physically} and adapt their attack loss into the two-modality form like our fitness evaluation to perform a comparison. As Figure \ref{fig:compare} (d) shows, despite its competitive naturalness with \cite{wei2022hotcold} and \cite{fischler1981random}, it still has slight sawtooth on the edge. As for the quantitative attack performance of the above methods, Table \ref{tab:compare} significantly shows our method's best performance among the four, achieving an ASR of 83.33\%, an AP drop of 94.31\% in the visible modality and 88.30\% in the infrared modality, while the other three have poorer attack performances obviously. 

% \begin{figure}[!h]
%     \centering
%     \includegraphics[width=0.8\linewidth]{shape_com.pdf}
%     \caption{A visualization  comparison for the optimized shape between \cite{chen2022shape} and ours.}
%     \label{fig:compare1}
% %\vspace{-0.3cm}
% \end{figure}

\begin{table*}[!htb]
\centering
   \caption{Attack performances against different object detection systems.}
\begin{tabular}{c |c| c | c | c | c | c }  \hline  
Detectors        & YOLOv3 & YOLOv5 & YOLOv7 & Faster RCNN  & EfficientDet & RetinaNet  \\   \hline  
ASR     &    83.33\%    &       71.67\%  &  80.00\%  & 69.17\%   &      67.50\% &   68.33\%             \\   \hline  
AP drop (Visible) &   94.31\%   &     82.11\%    &   89.66\%   &  89.38\%      &   76.48\%       &    82.11\%  \\     
AP drop (Infrared) &     88.30\%   &    87.18\%   &   93.84\%    &   83.94\%   &   84.77\% &     71.06\%  \\ \hline  
\end{tabular} 
\label{different detectors}
\end{table*}

\subsubsection{Attack Performances against Various Object Detectors}
\label{pdds}
Considering the differences between detection systems, we first verify our method in six typical and mainstream detectors: YOLOv3, YOLOv5, YOLOv7, Faster RCNN, EfficientDet and RetinaNet. We use ASR and AP drop to evaluate the attack performance. The results are shown in Table \ref{different detectors}.

From the above results, we can see that our method is equally useful despite the distinctions between the one-stage and two-stage detection models. For the typical one-stage detection models of the YOLO series, we first achieve an ASR of 83.33\%, an AP drop of 94.31\% in the visible modality and 88.30\% in the infrared modality for the classic model YOLOv3, then for the upgraded detectors YOLOv5 and YOLOv7, we still achieve good performances. In the meantime, for the two-stage detection model Faster RCNN, we achieve an ASR of 69.17\%, an AP drop of 89.38\% in the visible modality and 83.94\% in the infrared modality. As for EfficientDet and RetinaNet, our method can achieve similar results to Faster RCNN on these high-performance detectors, which confirms our method’s generality across different detection systems.

\begin{figure}[!h]
    \centering
    \includegraphics[width=0.98\linewidth]{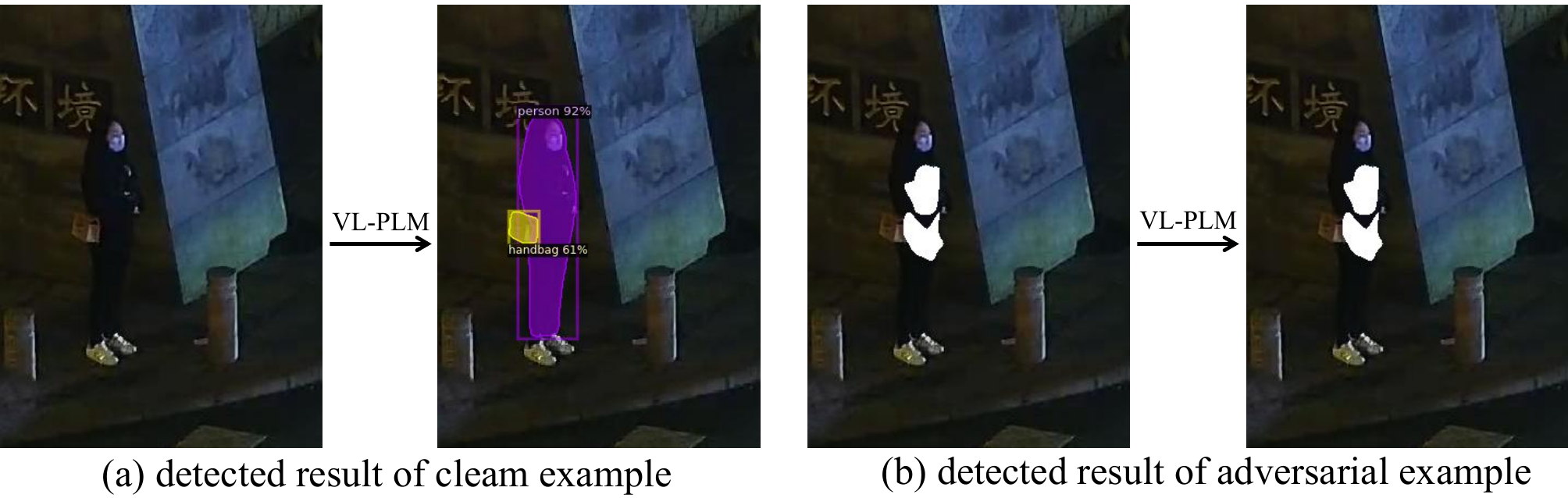}
    \caption{A visual example of attacks against CLIP-based detector.}
    \label{fig:clip}
\end{figure}

Then, we also try to test our attack against the latest CLIP-based detectors, like VL-PLM\cite{zhao2022exploiting}. For the target modality, due to CLIP-based detectors' custom usage in the visible modality, we follow it and select the visible light domain. For the attack method, because our method is a score-based black-box attack, faced with the single modality, we can also use the confidence scores to perform single-modality attacks. The corresponding ASR is 75.83\%, which shows that the current visual-language models are also vulnerable to our method. Figure \ref{fig:clip} shows the visualized attack effects on VL-PLM. Compared the two examples, we can observe that with our adversarial patches, this model with high-precision target detection (and even segmentation) capabilities has also lost its powerful function.

% As described in Section \ref{combined with position}, this phenomenon may be accounted for that our patches can cover sensitive areas of the object at most times as they move towards different regions. Results of gradcam\cite{Selvaraju_2019} for our samples in Figure \ref{fig:grad_cam} can verify our explanation, which shows that taking YOLOv5 as an example, the significant part of the pedestrian class that this detector focuses on is concentrated in the torso part of the human body and matches the initial position and the size of our unified adversarial patch to a large extent. 

%  \begin{figure}[ht]
% \begin{center}
% \includegraphics[width=\linewidth]{grad_cam.pdf}
% \end{center}
%    \caption{Significant part of the pedestrian class that YOLOv5 detector focuses on visualized by gradcam}
% \label{fig:grad_cam}
% \end{figure}
\subsubsection{Robustness to Implementation Errors}
\label{robustness}
Since the patch is in general generated digitally by considering the optimal shape and fixed location, when it is applied to real scenarios, it is natural to see how sensitive it would be to the attack success rate if we do not cut and paste the patch in 100\% exact shape and location on the clothes. Therefore, we simulate possible position shifts and clipping errors when conducting physical experiments. Table \ref{errors} demonstrates our cross-modal patches' robustness to translation errors and incompleteness.

\begin{table}[!htb]
\caption{ASR of simulating implementation errors against YOLOv3.}
\resizebox{\linewidth}{!}
{
\begin{tabular}{c|cc|cc}
\hline
\multirow{2}{*}{} & \multicolumn{2}{c|}{Translation} & \multicolumn{2}{c}{Incompleteness}
\\ \cline{2-5}  & \multicolumn{1}{c|}{3pix}  & 5pix  & \multicolumn{1}{c|}{5\%}  & 10\%  \\ \hline
ASR  & \multicolumn{1}{c|}{\begin{tabular}[c]{@{}c@{}}64.17\%\\ ($\downarrow$9.16\%)\end{tabular}} & \begin{tabular}[c]{@{}c@{}}55.00\%\\ ($\downarrow$18.33\%)\end{tabular} & \multicolumn{1}{c|}{\begin{tabular}[c]{@{}c@{}}68.83\%\\ ($\downarrow$5.00\%)\end{tabular}} & \begin{tabular}[c]{@{}c@{}}60.83\%\\ ($\downarrow$12.50\%)\end{tabular} \\ \hline
\end{tabular}
}
\label{errors}
\end{table}

\begin{figure*}[htp]
\begin{center}
\includegraphics[width=\linewidth]{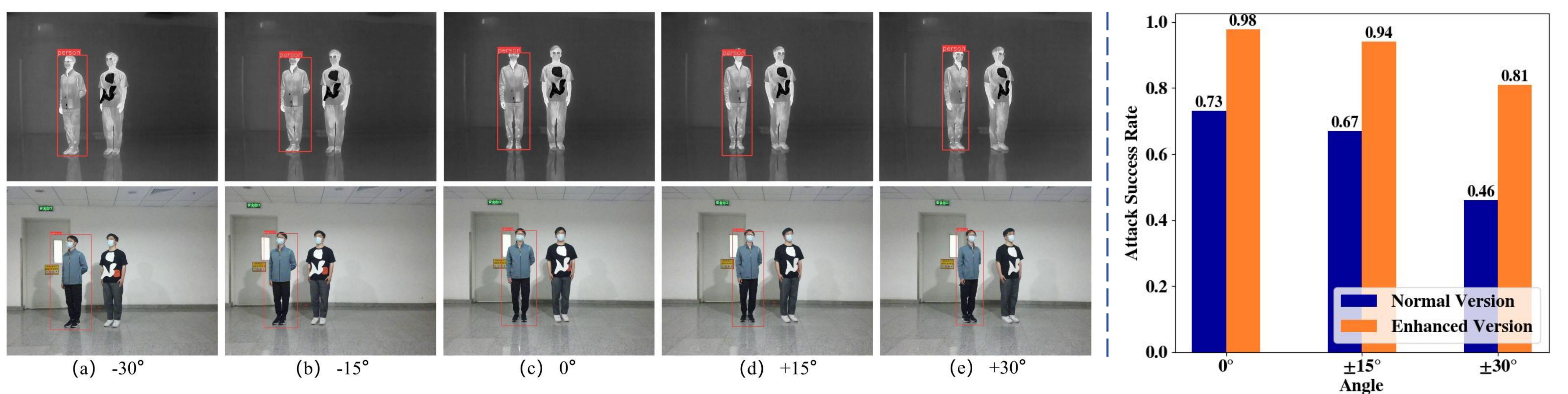}
\end{center}
\vspace{-0.5cm}
\caption{Qualitative and quantitative results of physical attacks with Affine-Transformation-based enhancement strategy under different angles. }
\label{fig:physical_angle}
\end{figure*}

\subsection{Attacks in the physical world}
\subsubsection{Physical Attacks against Pedestrian Detector}
To verify the effectiveness of unified adversarial patches in various physical settings, we design numerous scenarios for conducting physical attacks and capture videos to compute the ASR. By default, we make recordings of interior scenes at a distance of 4 meters from a standing person in the frontal view ($0^{\circ}$). For each situation, we shoot for 20 seconds at ten frames per second (about 200 frames in total) and compute the ASR versus the captured video. The pedestrian detection threshold is set at 0.7 according to \cite{zhu2022infrared}. 

For the angle problem, we change the angles with $\pm15^{\circ}$, $\pm30^{\circ}$. For distance, we move the camera to 6 meters from the default 4 meters. For posture, we change the pedestrian's posture from the default standing pose to the sitting pose. As for the scene, we change to outdoor from the default indoor. The visual examples of these situations are listed in Figure \ref{fig:physical}. During the shooting process, the pedestrian is asked to move the body within the range of $5^{\circ}$ of the current posture to take videos. 
Table \ref{tab:situations} lists the quantitative results. It can be seen  that our infrared patches achieve a high ASR (73.50\%) in the frontal view. When the shooting angle changes, the ASR still maintains a high value (67.00\% and 46.50\%). When changing the distance from 4 meters to 6 meters, the ASR decreases to 71.00\%.  When changing the posture from standing to sitting, the ASR decreases to 61.50\%, and ASR decreases to 57.00\% when the scene is changed to outdoor. These results show that the impact of different shooting situations is relatively small to the cross-modal patches. In other words, as long as the shape of our cross-modal patch on the object can be generally captured by the camera, the effects of adversarial attacks can be maintained. 

\begin{table}[!htb]
\caption{ASR in the physical world when changing angles, distances, postures and scenes captured by multi-modal sensors against YOLOv3.}
  \begin{center}
  \resizebox{\linewidth}{!}{
    \begin{tabular}{c|c|c|c|c|c|c}
    \hline
  Setting  &  $0^{\circ}$ &  $\pm 15^{\circ}$ & $\pm 30^{\circ}$  & dist. & pos. & outdoor \\
    \hline
    ASR & 73.50\% & 67.00\%  & 46.50\% & 71.00\% & 61.50\% &57.00\%\\
    \hline
    \end{tabular}}
    \label{tab:situations}
  \end{center}
\end{table}

\subsubsection{Affine-Transformation-based Enhancement Strategy}
\label{AT-ES}
Considering that pedestrians always have different angular deflections under object detectors when moving, we aim to generate more robust shapes that can be fully detected at multiple angles with a stronger adversarial effect. Specifically, we use the multi-angle joint optimization method proposed in Section \ref{multi-angles} to obtain the optimal patches' shapes that are more applicable for multi-angle situations, and the visualization results of the attack performances compared with the clean target is shown in Figure \ref{fig:physical_angle}. Then, we contrast the enhanced version with the normal version and the quantitative results are also given in Figure \ref{fig:physical_angle}, where we can see that with the proposed enhancement strategy, the attack success rate has generally increased by a significant average ASR of 30\%. Besides, compared with Figure \ref{fig:physical}, we can intuitively observe that patch generation that takes angle factor into account may tend to learn a bilateral distribution as shown in Figure \ref{fig:physical_angle} rather than a relatively centralized one. This could be interpreted by the fact that patches located in the middle area of the object may be detected as a narrow strip at a large deflection angle, which will lose attack performance in the physical world.

\subsubsection{Physical Attacks against Vehicle Detector}
\label{PA-VD}
\begin{figure}[!h]
    \centering
    \includegraphics[width=\linewidth]{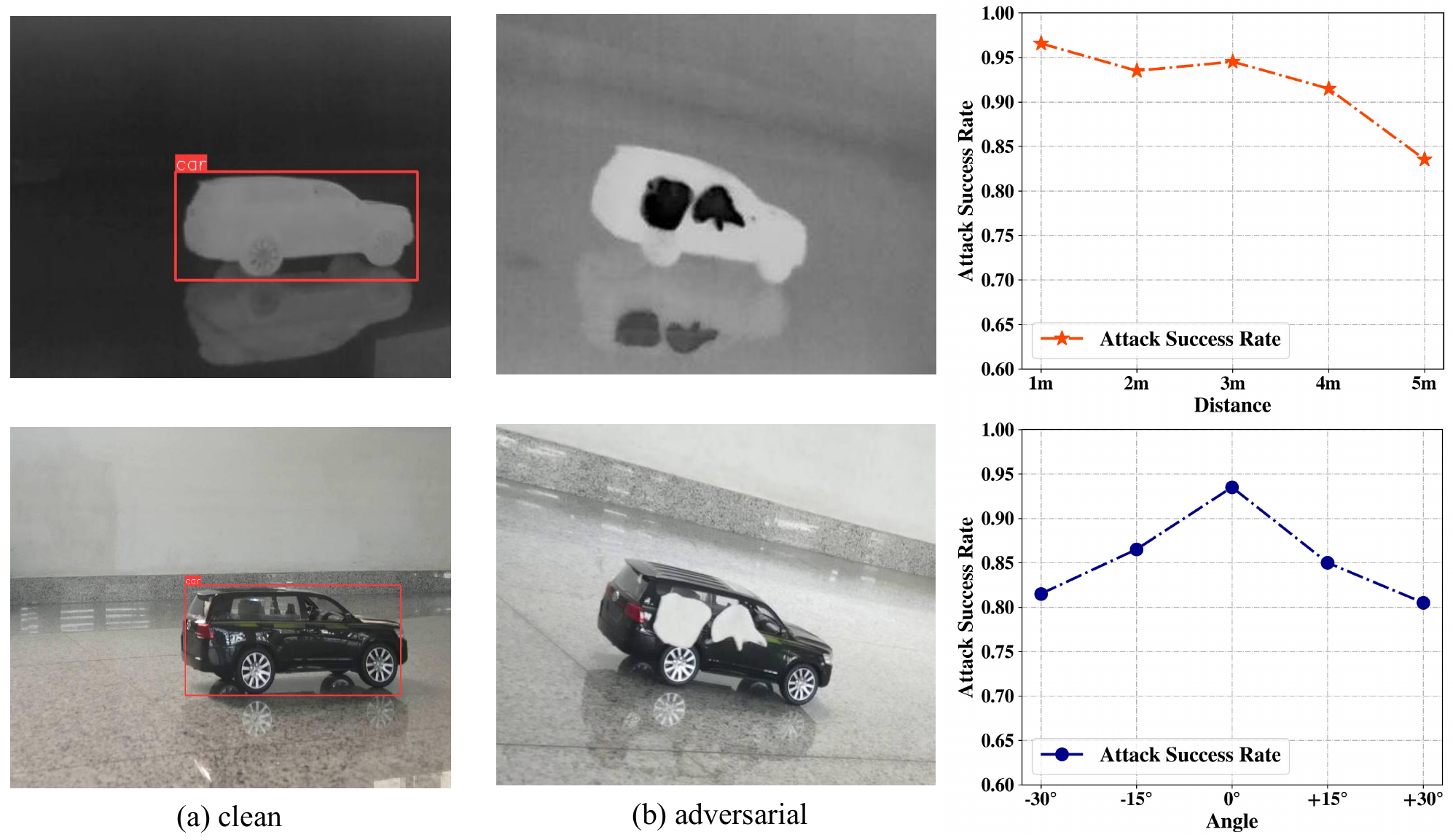}
    \caption{Qualitative and quantitative results of physical attacks against vehicle detector under different distances and angles.}
    \label{fig:vehicle}
\end{figure}
Besides the pedestrian detection task, our unified adversarial patch can also be applied to other tasks. In this section, we conduct the experiments of physical attacks against vehicle detection based on YOLOv3. Like the pedestrian detection task, we record the vehicle video (200 frames per distance) captured at different distances (1m, 2m, 3m, 4m, 5m) and different angles(-30$^\circ$, -15$^\circ$, 0$^\circ$, +15$^\circ$, +30$^\circ$), and then compute the ratio of successfully attacked video frames out of all the frames as the ASR.

The specific results are given in Figure \ref{fig:vehicle}, where the left denotes the qualitative examples and the right denotes the quantitative ASR against different distances and angles. To give a clear comparison, we list two vehicle images of the same modality in one row, where the vehicle attached by the unified adversarial patch cannot be detected but the clean vehicle is successfully detected. Additionally, we see the patch' size is small versus the vehicle, but it indeed performs successful adversarial attacks, which proves the vulnerability of vehicle detector against the unified adversarial patch. The right quantitative results show the high attack performance versus different distances (96.50\% at 1m, 93.50\% at 2m, 94.50\% at 3m, 91.50\% at 4m, 83.50\% at 5m) and different angles (81.5\% at -30$^\circ$, 86.50\% at -15$^\circ$, 93.50\% at 0$^\circ$, 85.00\% at +15$^\circ$, 80.50\% at +30$^\circ$). It shows the similar trends with the pedestrian detection task.

\subsection{Defenses against  Unified Adversarial Patches}
\label{defense}
\begin{table}[!h]
\caption{Results against the defense methods.}
  \begin{center}
    \begin{tabular}{c|c|c}
    \hline
    Defense Methods &  ASR  & Error\\
    \hline
    No Defense   & 73.33\% &0.00\% \\
    \hline
    Spatial Smoothing \cite{xu2017feature} & 64.17\% &9.16\%\\
    \hline
    Adversarial Training \cite{goodfellow2014explaining} & 50.00\% & 23.33\%\\
    \hline
    \end{tabular}
    \label{tab:defense1}
  \end{center}
\end{table}
We test two typical methods to defend our attack method in the digital world. One is pre-processing defense: spatial smoothing \cite{xu2020adversarial}, and another is adversarial training \cite{goodfellow2014explaining}.  The defense results are given in Table \ref{tab:defense1}, where we see that: (1) After spatial smoothing, the ASR only drops 9.16\%. This is reasonable because the cover image in our method has the same value, the smooth operation cannot change the distribution. (2) After adversarial training, the ASR drops 23.33\%, which is still acceptable. It shows the robustness of our unified adversarial patches.

\section{Conclusion}

In this paper, we proposed a unified adversarial patch in the physical world. For that, we uncovered the property that could react both in the visible and infrared modalities: shape. Then, combining the boundary-limited shape optimization with the score-aware iterative fitness evaluation, we guaranteed an efficient exploration of the adversarial shape and the balance between different modalities. We also proposed an Affine-Transformation-based enhancement strategy to cope with the shape deformation caused by different shooting angles in the real world. Experiments on the pedestrian detection and vehicle detection tasks in the digital world and physical world verified the effectiveness of our proposed method. In the future, we will apply our method on more visual tasks to demonstrate its good generalization.

\ifCLASSOPTIONcaptionsoff
  \newpage
\fi

\bibliographystyle{IEEEtran}
\bibliography{egbib}

\end{document}